\documentclass[pdflatex,sn-mathphys-num]{sn-jnl}
\usepackage{svg}

\usepackage{graphicx}%
\usepackage{multirow}%
\usepackage{amsmath,amssymb,amsfonts}%
\usepackage{amsthm}%
\usepackage{mathrsfs}%
\usepackage[title]{appendix}%
\usepackage{xcolor}%
\usepackage{textcomp}%
\usepackage{manyfoot}%
\usepackage{booktabs}%
\usepackage{algorithm}%
\usepackage{algorithmicx}%
\usepackage{algpseudocode}%
\usepackage{listings}%
\usepackage{placeins} 

\usepackage{adjustbox}
\usepackage{tabularx}

\usepackage[utf8]{inputenc}
\usepackage{tabularx}
\usepackage{booktabs}
\usepackage{array}
\usepackage{ragged2e}

\usepackage{colortbl}

\theoremstyle{thmstyleone}%
\theoremstyle{thmstyletwo}%

\theoremstyle{thmstylethree}%

\begin{document}


\title[Article Title]{A document processing pipeline for the construction of a dataset for topic modeling based on the judgments of the Italian Supreme Court}


\author*[1]{ \sur{Matteo Marulli}}\email{matteo.marulli@unifi.it}

\author[2]{ \sur{Glauco Panattoni }}\email{glauco.panattoni@unifi.it}

\author[1]{ \sur{Marco Bertini}}\email{marco.bertini@unifi.it}

\affil[1]{\orgdiv{Dipartimento di ingegneria dell'informazione}, \orgname{Università degli studi di Firenze}, \orgaddress{\street{ Viale Giovanni Battista Morgagni, 65, 50134 Florence FI}, \country{Italy}}}

\affil[2]{\orgdiv{Dipartimento di Scienze Giuridiche}, \orgname{Università degli studi di Firenze}, \orgaddress{\street{Via delle Pandette, 32, 50127 Firenze FI}, \city{Florence}, \country{Italy}}}

\affil[3]{\orgdiv{Dipartimento di ingegneria dell'informazione}, \orgname{Università degli studi di Firenze}, \orgaddress{\city{ Viale Giovanni Battista Morgagni, 65, 50134 Florence FI}, \country{Italy}}}


\abstract{Topic modeling in the context of Italian jurisprudence is limited by the lack of publicly available datasets, which hampers the analysis of legal themes covered by the Italian Supreme Court. To overcome this challenge, we developed a document processing pipeline that generates an anonymized dataset of legal judgments optimized for topic modeling.

The pipeline combines document layout analysis, optical character recognition, and text anonymization. The DLA module, based on YOLOv8x, achieved a mAP@50 of 0.964 and a mAP@50–95 of 0.800. The OCR text detector based on YOLOv8x reached a mAP@50–95 of 0.9022, while the text recognizer (TrOCR) obtained a Character Error Rate (CER) of 0.0047  and a Word Error Rate (WER) of 0.0248. The dataset enabled topic modeling with a diversity score of 0.6198 and a coherence score of 0.6638, surpassing traditional OCR-only approaches.

We applied BERTopic to extract topics and used Large Language Models (LLMs) to generate labels and summaries. Outputs were evaluated against domain expert interpretations, showing comparable semantic quality. Specifically, Claude Sonnet 3.7 achieved a BERTScore F1 of 0.8119 for topic labeling and 0.9130 for topic summarization.}

\keywords{document-processing, topic-modeling, computer-vision, large-language
models-(LLMs), Italian-legal-documents}



\maketitle

\section{Introduction}\label{sec1}

Supervised learning techniques \cite{hastie2009overview}, such as neural networks \cite{8511194}, support vector machines (SVMs) \cite{clavie2021unreasonable}, and decision trees (DTs) \cite{thammaboosadee2012framework}, are widely applied in legal AI for classification and regression tasks, making them essential for applications such as legal document categorization and case outcome prediction. In contrast, unsupervised methods \cite{solorio2020review}, including clustering \cite{prince2023exploring} and topic modeling \cite{abdelrazek2023topic}, facilitate exploratory data analysis by uncovering latent structures in large textual datasets, such as legal topic discovery. However, their adoption remains limited in the NLP context for the Italian legal domain due to the scarcity of datasets explicitly designed for exploratory purposes and the need for expert involvement to contextualize results.

Legal topic modeling datasets for Italian case law are particularly rare, making it difficult to identify recurring themes in legal documents, extract knowledge, and analyze large volumes of normative and jurisprudential texts. This limitation hinders the development of advanced tools for information retrieval and automated judgment categorization, increasing reliance on manual analysis and expert work.

Furthermore, legal datasets contain sensitive information about individuals or companies, imposing restrictions on their sharing due to privacy and data protection concerns. Limited access to these resources hampers the creation of open benchmarks and the reproducibility of studies, slowing progress in the application of topic modeling to legal texts. Additional constraints arise from complex data protection regulations, such as the GDPR (General Data Protection Regulation) \cite{voigt2017eu} in Europe, which requires anonymization strategies to safeguard privacy

Building a dataset is a complex and costly process that requires significant investment in time and resources. Computer vision emerges as a promising approach for automatically generating datasets for natural language processing (NLP) applications, leveraging advanced methods for analyzing and processing textual documents. Optical character recognition (OCR) and document layout analysis (DLA) techniques \cite{subramani2020survey},\cite{binmakhashen2019document} are examples of such methods.

To address these challenges, we propose a document processing pipeline to create a dataset for topic modeling using rulings from the Italian Supreme Court\footnote{Authors’ note: In this contribution, we refer to the Italian Supreme Court as the Corte di Cassazione. It is the highest judicial body in civil and criminal matters in the Italian legal system.} as a source.

Our pipeline initially employs computer vision methods to analyze the structure of judgments, enabling the accurate extraction and segmentation of textual content. This step improves OCR accuracy and ensures that the extracted text is properly structured for subsequent NLP tasks.

To ensure privacy compliance, we integrate a transformer-based Named Entity Recognition (NER) model capable of detecting and anonymizing sensitive information, such as personal names, addresses, emails, tax codes, and dates, in line with GDPR guidelines.

After constructing the dataset, we apply topic modeling to identify key topics. Subsequently, we employ large language models (LLMs) \cite{zhao2023survey} to interpret the extracted topics by generating summaries and labels for each. However, due to their inherent biases and lack of legal expertise, we will systematically compare LLM-generated interpretations with those provided by legal experts, ensuring a critical assessment of their reliability in a judicial context.

In this paper, we introduce a document processing pipeline that combines computer vision and NLP algorithms to create a novel dataset for topic modeling in the Italian jurisprudence domain. Our main contributions are:

\begin{itemize}

  \item The construction of a new dataset of Italian Supreme Court judgments through document processing techniques and its impact in topic modeling.

  \item The application of topic modeling in the legal domain and in the Italian language.

  \item The application of LLMs to interpret topics by generating labels and summaries describing the topics and comparing them with the labels and summaries generated by a domain expert.

\end{itemize}

\section{Related Work}\label{sec2}

\subsection{Supervised Learning in Legal NLP}
Supervised machine learning has been a key pillar in the development of NLP applications for the legal domain, with significant results in the classification of legal documents, although its effectiveness depends critically on the availability of reliable ground truth. Undavia et al. \cite{8511194} conducted a comparative study on the application of neural networks for the classification of U.S. Supreme Court opinions using the University School of Law Supreme Court Database (SCDB) with manual labels created by legal experts, demonstrating that a CNN combined with pre-trained word embeddings achieved 72.4\% accuracy in classifying into 15 broad legal categories, a result only possible due to the quality of manual annotation created by legal specialists.

Despite the advancement of complex architectures, recent research has highlighted a counterintuitive phenomenon in the legal domain that underscores the importance of ground truth beyond algorithmic complexity. Clavié and Alphonsus \cite{clavie2021unreasonable} have shown, using the LexGLUE benchmark with expert labeling, that traditional SVM classifiers perform surprisingly competitively against BERT models, with an error reduction of only 18.1\% in the legal domain versus 85.1\% in the general domains. In parallel, Thammaboosadee et al. \cite{thammaboosadee2012framework} have proposed a two-stage approach for identifying applicable law items, highlighting how the quality of legal annotations at different levels of abstraction (facts, diagnostic problems, legal elements) is crucial to the success of supervised learning in this domain.

These studies converge on one essential point: while supervised learning offers promising results in legal NLP, its applicability is highly constrained by the availability of datasets with ground truth created by legal experts, an expensive and scarce resource, especially in languages other than English and in less studied jurisdictions. In our work, we acknowledge these limitations and propose to explore the application of unsupervised topic modeling to Italian jurisprudence, an approach that, although it does not require manual labeling for training, will still benefit from expert interpretation for validating results, thus establishing a bridge between supervised and unsupervised methods for analyzing legal texts.

\subsection{Unsupervised Learning and Topic Modeling}

Unsupervised learning techniques, particularly clustering and topic modeling, are gaining increasing interest in the legal NLP field due to their ability to analyze large amounts of textual data without the need for manual annotation. However, their use in the context of the Italian language and specifically in the legal field remains limited.

Recently, Giaconia et al. \cite{giaconia2024topic} applied topic modeling techniques such as LDA, ETM and ProdLDA to Italian bank reports for the identification of suspicious AML-related activities. Although the results demonstrate the usefulness of such techniques in detecting hidden patterns and latent semantics in bank texts, the authors do not provide in-depth details regarding the data preparation and acquisition phase, leaving it uncertain whether they used manual transcription techniques or automatic computer vision-based methods.

The work of Cataldo et al. \cite{cataldo2021divorce} represents a contribution in the area of textual analysis of Italian legal judgments, as it applies the Latent Dirichlet Allocation (LDA) model to extract latent themes from Supreme Court judgments concerning divorce. This approach identified three main dimensions: the procedural stages of divorce, difficulties in the transition from separation to divorce, and socio-economic measures related to post-divorce maintenance. However, as in Giaconia et al.'s work, the authors do not explain how they prepared the documents for subsequent processing because judgments are long documents and most machine learning or deep learning models struggle to process fairly long sequences of data. When working with long documents, it is advisable to segment them. In this way, the content of each document can be fully exploited.

In parallel, work such as that of Vianna et al. \cite{vianna2024topic} and Silveira et al. \cite{silveira2021topic}, although applied to Portuguese and English legal texts respectively, have highlighted the advantages of advanced techniques such as ETM, CTM, BERTopic, and legal domain-specific embeddings (e.g., LEGAL-BERT). Both studies highlight the importance of incorporating specific legal context to improve the semantic quality of textual representations, although both work with texts in languages other than Italian.

With regard specifically to the Italian domain, the scarcity of freely available datasets is often attributable to constraints related to privacy and the protection of sensitive data. In addition, little information is usually provided on the methodology adopted for the preparation of the datasets themselves: for example, it is not made clear whether the texts of the judgments were transcribed manually, extracted automatically through OCR, or through more sophisticated computer vision-based document structure analysis techniques.

\subsection{Computer Vision for Document Processing}

The integration of computer vision techniques in NLP tasks has significantly improved the extraction and structuring of textual information from complex documents. Optical Character Recognition (OCR) and Document Layout Analysis (DLA) are essential components in preprocessing workflows that enable the accurate segmentation and organization of legal texts for downstream NLP applications.

Despite their success in other domains, DLA and OCR techniques have been relatively underexplored in the processing of legal documents. The highly structured nature of legal texts, presents unique challenges that have yet to be fully addressed by existing methodologies.

Document Layout Analysis (DLA) is particularly crucial for handling the heterogeneous structure of legal texts, which often contain footnotes, citations, and multiple sections. BinMakhashen and Mahmoud \cite{binmakhashen2019document} provide a comprehensive survey of DLA techniques, outlining preprocessing steps, segmentation strategies, and evaluation metrics. They highlight that the diversity of document layouts necessitates robust segmentation strategies, including bottom-up, top-down, and hybrid approaches, to effectively partition legal documents into meaningful sections.

In the context of OCR, recent work by Rang et al. \cite{rang2023empirical} investigates scaling laws for large-scale OCR models, demonstrating that larger models benefit from increased data volume and computational resources. This research supports the premise that high-quality OCR pipelines require not only sophisticated recognition architectures but also extensive training data tailored to the legal domain.

Audebert et al. \cite{audebert2019multimodal} propose a multimodal approach that fuses OCR-extracted text embeddings with image-based features to enhance document classification tasks. Their findings indicate that combining textual and visual cues significantly improves classification accuracy, a strategy that can be leveraged for legal document processing, particularly in structuring rulings for topic modeling.

Another study by Baimakhanova et al. \cite{baimakhanova2024multimodal} applies deep learning to document categorization, emphasizing the role of segmentation in handling scanned legal documents. They employ convolutional neural networks (CNNs) to extract structural patterns, reinforcing the necessity of automated preprocessing for efficient document management.

In the domain of clinical information extraction, Hsu et al. \cite{hsu2022deep} present a deep learning pipeline that integrates image preprocessing and OCR to extract key information from scanned medical reports. Their approach, which combines NLP with structured document analysis, underscores the importance of preserving layout information when processing scanned legal texts.

\subsection{Named Entity Recognition for Privacy Compliance}
Privacy compliance in the context of legal textual data is critically important due to regulatory frameworks like the GDPR, which impose stringent requirements on the handling and dissemination of sensitive personal information. Recent research has increasingly focused on employing advanced Natural Language Processing (NLP) techniques, particularly Named Entity Recognition (NER), to automate the anonymization process

Licari et al. \cite{licari2022automatic} presented one of the earliest systems specifically designed for automatic anonymization in the Italian legal domain. Utilizing Transformer-based models and spaCy’s transition-based parsing, their system effectively identified and anonymized sensitive entities achieving more than 94.7\% recall, notably reaching greater than 99\% recall for personal names and identification entities. This system marked a significant advancement in automating GDPR-compliant anonymization of judicial documents.

GiusBERTo \cite{salierno2024giusberto} further advanced this direction by introducing a BERT-based model specifically fine-tuned for anonymizing personal data within decisions of the Italian Court of Auditors. The model was trained on an extensive corpus of judicial documents, achieving a high level of token-level accuracy (97\%), demonstrating robust performance in balancing data protection and contextual relevance.

In a complementary direction, GLiNER \cite{zaratiana2023gliner} offers a generalized approach to entity recognition capable of recognizing arbitrary entity types using a bidirectional transformer. GLiNER stands out due to its ability to perform parallel entity extraction efficiently, demonstrating strong zero-shot generalization across diverse NER benchmarks. Such generalization capabilities are crucial for addressing anonymization tasks across various document types and legal sub-domains, reducing dependence on specific training datasets.

\subsection{Large Language Models in Legal NLP}
The interpretation of topics generated by traditional topic modeling models, such as Latent Dirichlet Allocation (LDA) and BERTopic, is an open challenge in the field of text analysis. Representations based on word distributions are often difficult to understand, requiring experts to assign consistent meanings to extracted topics. Recent studies have explored the use of Large Language Models (LLMs) to fill this gap, taking advantage of their ability to generate readable and contextualized text descriptions. TopicGPT \cite{pham2024topicgpt} introduced a framework that uses GPT-4 to assign semantic labels to topics more intuitively than traditional methods. This approach demonstrated greater consistency with human categorizations, outperforming LDA in semantic alignment metrics and cluster purity, and reducing ambiguity in the interpretation of results.

In parallel, other studies have investigated the role of LLMs in explaining topics generated by traditional models. One significant experiment used ChatGPT to generate textual descriptions from LDA topics applied to medical data \cite{rijcken2023towards}. Comparison of ChatGPT descriptions with those provided by experts showed that about 50 percent of the interpretations generated by the model were considered useful, suggesting that LLMs can indeed assist in topic understanding if guided by appropriate prompts. However, the risk of incorrect or overly general interpretations highlights the need for human control, making these tools a complement rather than a substitute for manual analysis.

These studies show that LLMs can improve the accessibility and readability of topic modeling results, making exploratory analysis of large textual datasets easier. However, several challenges remain open, including the risk of bias and the need for expert validation to ensure the reliability of interpretations

\subsection{Summary and Positioning}
The literature review reveals several significant research gaps in the application of topic modeling to Italian legal texts. While supervised learning techniques have shown promising results in classifying legal documents, they remain heavily dependent on expert-annotated datasets, which are expensive and scarce. Unsupervised methods such as topic modeling offer a viable alternative that does not require labeled data however the construction of these datasets remains challenging in terms of time and resources, so their application to Italian jurisprudence has been limited, with existing studies providing insufficient detail on data preparation and document processing methodologies.

In addition, privacy and GDPR compliance concerns present significant challenges in creating sharable legal datasets. Although recent advances in NER have demonstrated effectiveness in anonymizing sensitive information in legal texts, the integration of such techniques within comprehensive document processing pipelines remains underexplored. Similarly, computer vision approaches for document layout analysis and OCR, despite their potential to accurately extract and structure legal texts, have not been systematically applied to Italian Supreme Court judgments.

Moreover, while LLMs show promise in the interpretation of topic modeling results, their application in the legal field raises questions about reliability and expert validation, particularly when dealing with specialized legal terminology and concepts in languages other than English.

Our research addresses these shortcomings by proposing an end-to-end pipeline that combines computer vision techniques for document analysis, OCR for text extraction, NER for privacy-compliant anonymization, and topic modeling for knowledge discovery from Italian Supreme Court judgments. Unlike previous studies, we provide detailed methodological insights into the dataset creation process, emphasizing reproducibility and transparency while maintaining privacy-compliance.

In addition, our work goes beyond simple topic extraction by critically evaluating LLMs' abilities to interpret legal topics through comparison with expert annotations. This approach not only contributes a new dataset, but also advances the understanding of the strengths and limitations of automated topic interpretation in specialized domains. By focusing on the Italian language and legal system, our research contributes to bridging the gap in multilingual legal NLP, offering insights and resources that could benefit similar efforts in other languages and jurisdictions with comparable structural challenges.

\section{Dataset}\label{AA}
\subsection{Data source} \label{sec:Data_source}

The Italian Supreme Court publishes its judgments, ordinances, and decrees annually on the Italgiure website\footnote{\url{https://www.italgiure.giustizia.it/sncass/}}, providing access to legal documents in PDF format. These documents are generally of high quality, as most scans are free from defects such as skewness, blurring, photometric alterations, compression artifacts, and low resolution. However, minor imperfections can occasionally be found. Handwritten scribbles or corrections may appear in some cases, while ink defects affecting entire pages or lines are rare but not entirely absent.

Given the extensive availability and high quality of these documents, we selected a subset of judgments for analysis. Specifically, we randomly chose 161 civil judgments and 146 criminal judgments, resulting in a total of 307 cases.

After assembling the corpus of judgments, we performed a quantitative analysis to examine its internal structure. As a first step, we assessed the overall length of the documents, uncovering significant differences between civil and criminal judgments.

\begin{figure}[htbp]
    \begin{minipage}{0.5\textwidth}
        \centering
        \includegraphics[width=\linewidth]{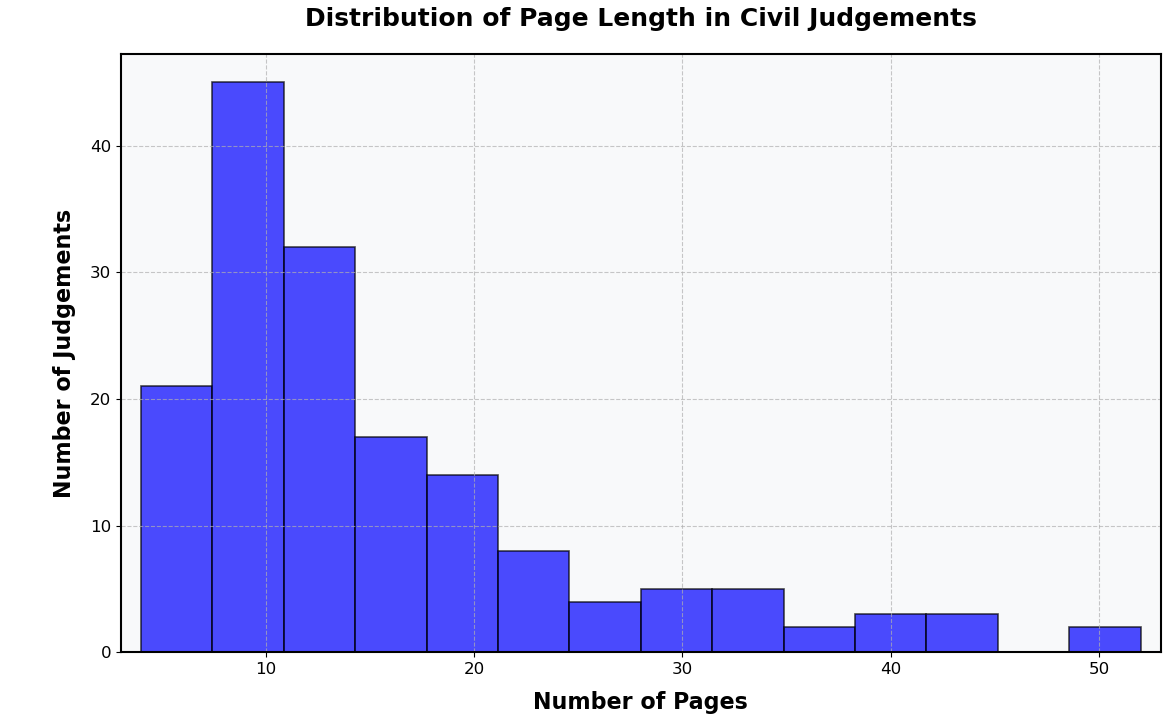}
        \caption{Distribution of page length of civil judgments of our dataset.}
        \label{fig:civil_dist}
    \end{minipage}
    \hfill
    \begin{minipage}{0.5\textwidth}
        \centering
        \includegraphics[width=\linewidth]{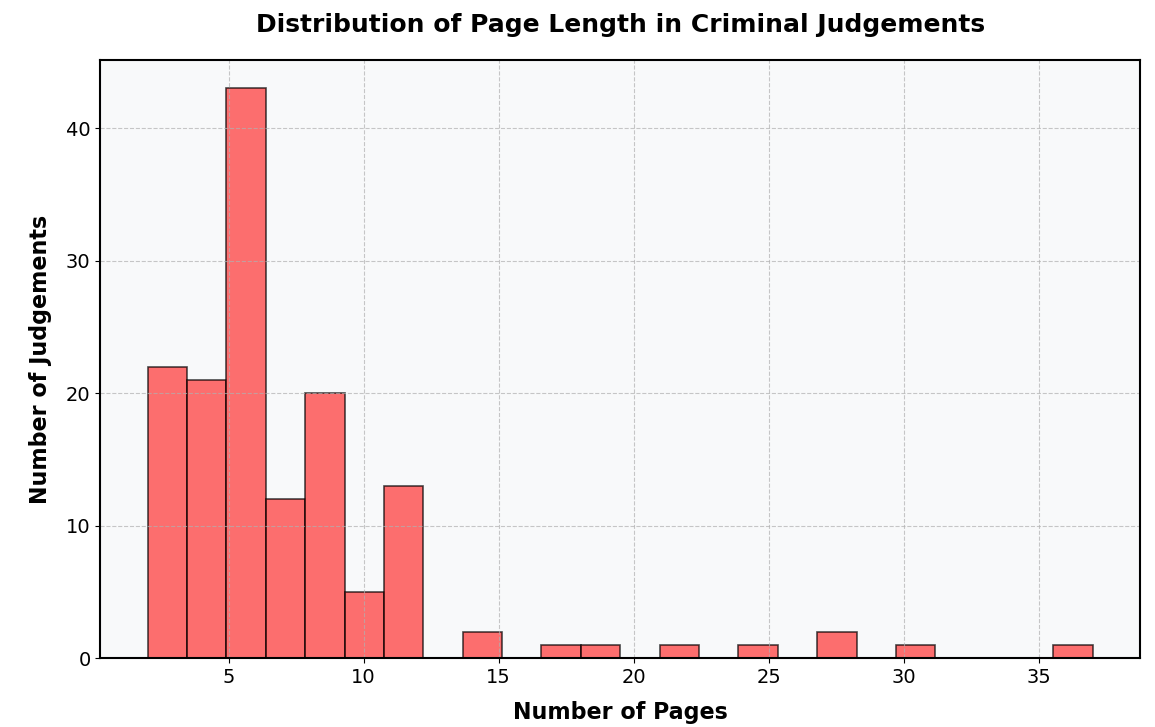}
        \caption{Distribution of page length of criminal judgments of our dataset.}
        \label{fig:criminal_dist}
    \end{minipage}
\end{figure}

Civil judgments tend to be longer than criminal judgments, as illustrated in Figure \ref{fig:civil_dist}. This difference is particularly evident in the distribution of page counts: civil judgments exhibit a longer right tail, indicating the presence of significantly lengthier cases. In contrast, the distribution of criminal judgments, shown in Figure \ref{fig:criminal_dist}, features a shorter and more dispersed right tail, suggesting greater variability in document length.

\begin{figure}
    \centering
    \includegraphics[width=0.85\linewidth]{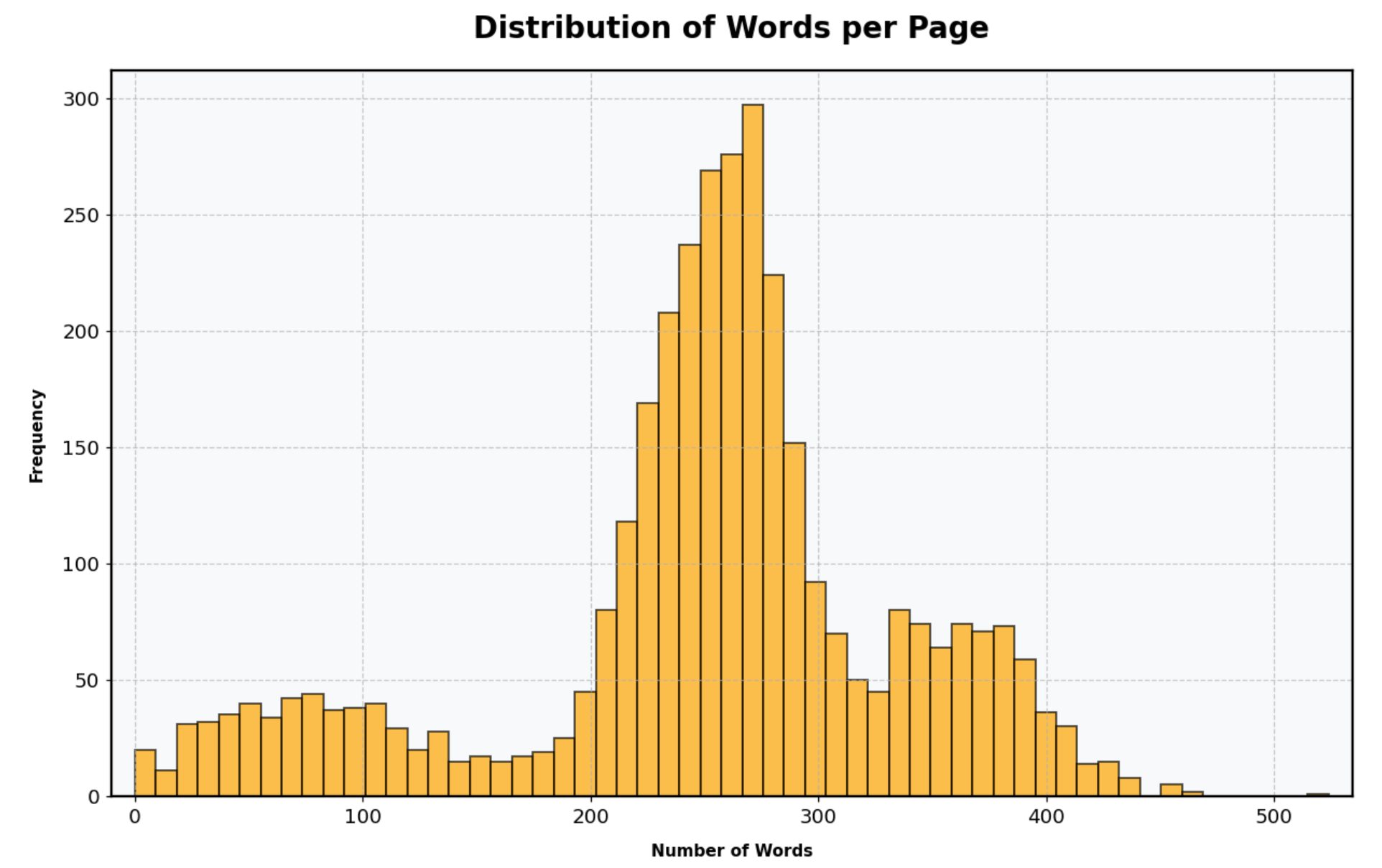}
    \caption{The distribution of the number of words per page of the judgments from our dataset.}
    \label{fig:distribution_of_words_per_page}
\end{figure}

The distribution of the number of words per page indicates that most sentences are densely packed with text, averaging around 250 words per page and occasionally exceeding 400, as shown in Figure \ref{fig:distribution_of_words_per_page}. This high word density poses a challenge for NLP models, as judgment pages contain extensive information that requires careful preprocessing. Without such preprocessing, models like BERT may struggle to fully leverage the content due to their maximum context length of 512 tokens.

While BERT's context length may seem sufficient to process the content of a single page, several factors complicate its practical application. Transformer models like BERT and GPT are primarily trained on English corpora and later adapted to other languages, including Italian. This adaptation affects tokenization, as BERT’s WordPiece algorithm \cite{wu2016google} constructs a vocabulary based on its training data. Consequently, many Italian words are split into multiple tokens if they are not present in the pre-built dictionary. As a result, numerous pages surpass the 512-token limit, causing truncation and potential loss of critical information, which may reside in the omitted portion. To address this issue, our work proposes a solution. In the following subsections, we describe the dataset annotation process.

\subsection{Dataset for document layout analysis}

To build a dataset for the Document Layout Analysis (DLA) task, modeled as an object detection problem, we downloaded an additional 59 judgments. We then annotated the layout of these documents using Label Studio \cite{labelstudio}, following the annotation scheme of the DocLayNet dataset \cite{pfitzmann2022doclaynet}. This structured approach enabled us to leverage transfer learning, transferring knowledge from DocLayNet to our dataset.

Ensuring accurate text segmentation was a crucial step. We verified that text-related bounding boxes aligned with paragraphs, as paragraph-level segmentation offers a balanced trade-off between processing efficiency and information completeness. Segmenting entire pages as text risks truncation due to BERT's 512-token limit, whereas finer segmentation at the line or sentence level can lead to overly fragmented text and increased computational costs.

The object classes we reused from DocLayNet are:
\begin{itemize}
\item Text: Regular paragraphs;
\item Section-header: Any kind of heading in the text, except the overall judgment title;
\item Page-footer: Repeating elements, such as page numbers at the bottom, outside of the normal text flow;
\item Title: The overall title of a judgment, (almost) exclusively on the first page and typically appearing in a large font.
\end{itemize}

\begin{figure}[htbp]
    \begin{minipage}{0.48\textwidth}
        \centering
        \includegraphics[width=\linewidth]{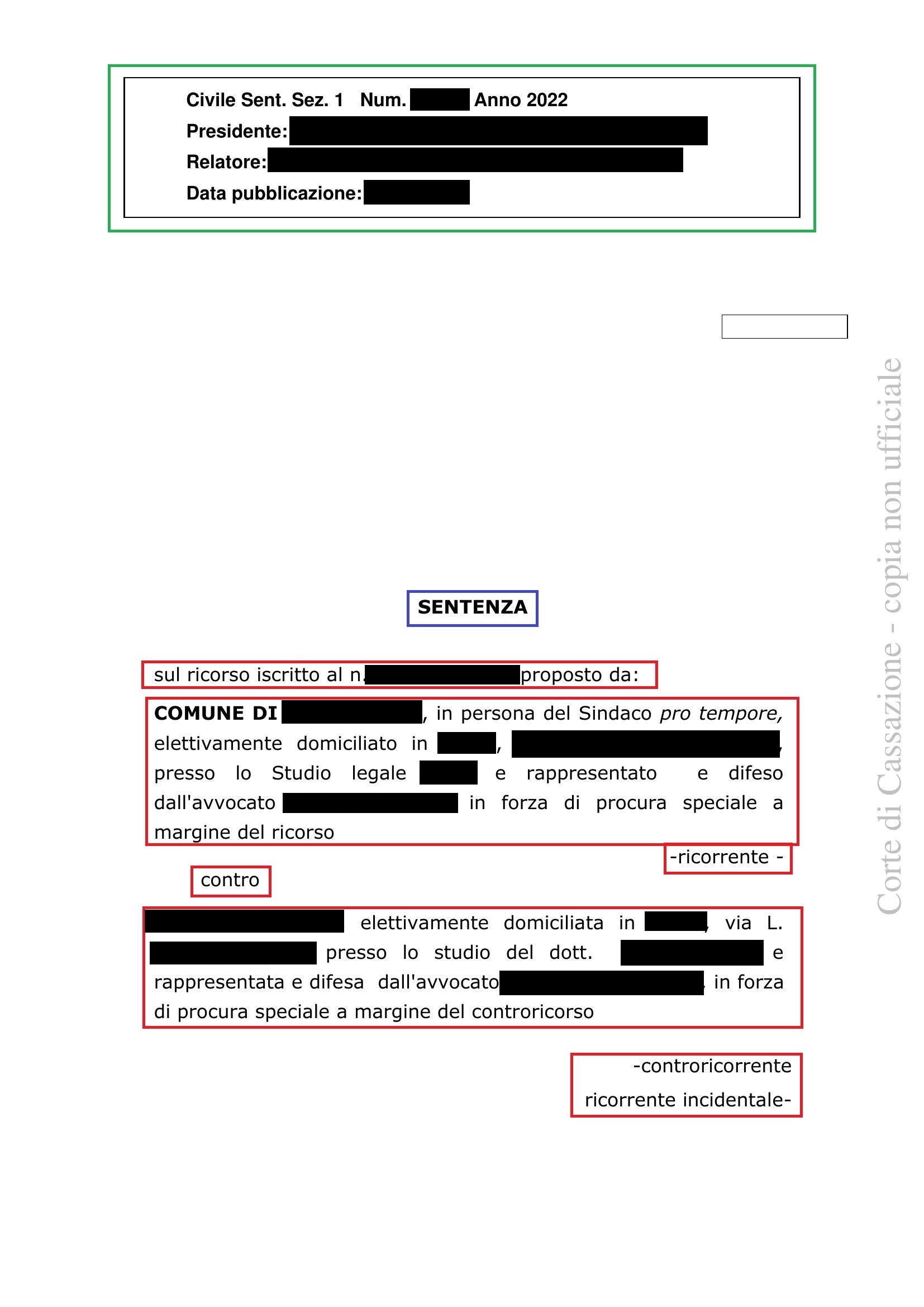}
        \caption{Front page of a annotated civil judgment of the Supreme Court.}
        \label{fig:civil_first_page}
    \end{minipage}
    \hfill
    \begin{minipage}{0.48\textwidth}
        \centering
        \includegraphics[width=\linewidth]{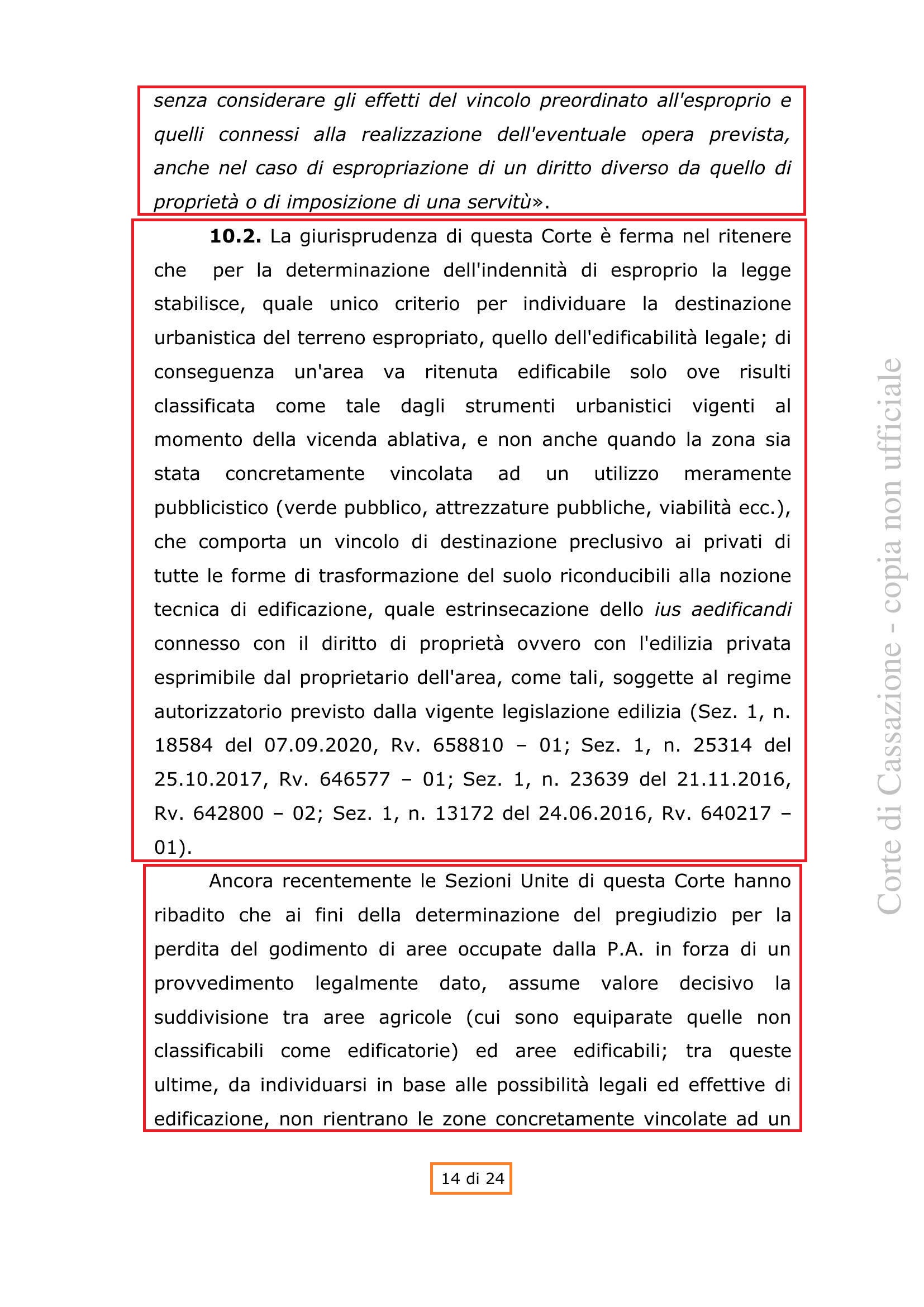}
        \caption{A page of the same annotated civil judgment}
        \label{fig:civil_last_page}
    \end{minipage}
\end{figure}

Figures \ref{fig:civil_first_page} and \ref{fig:civil_last_page} illustrate our document annotation approach. Paragraphs are enclosed in red bounding box, while section headers are marked as blue bounding box. Page footers are highlighted as orange bounding box, and title are distinguished by green bounding box.

We conclude this subsection by presenting Tables \ref{tab:dla_dataset_distributions} and \ref{tab:annotation_distribution_02}, which summarize key statistics of our dataset. Table \ref{tab:dla_dataset_distributions} reports the overall size of the image dataset, while Table \ref{tab:annotation_distribution_02} details the number of annotations per class along with their percentage frequencies, providing insights into the dataset's composition.

\begin{table}[h!]
\centering
\setlength{\tabcolsep}{6pt}
\renewcommand{\arraystretch}{1.2}
\begin{tabular}{l|ccc}
\hline
\textbf{Dataset Total} & \textbf{Training-set} & \textbf{Validation-set} & \textbf{Test-set} \\
\hline
799 & 634 (80\%) & 83 (10\%) & 82 (10\%) \\
\hline
\end{tabular}
\caption{Distribution of document images in our DLA dataset, split according to an 80-10-10 ratio.}
\label{tab:dla_dataset_distributions}
\end{table}

\begin{table}[h!]
\centering
\setlength{\tabcolsep}{8pt}
\renewcommand{\arraystretch}{1.2}
\begin{tabular}{l|rr}
\hline
\textbf{Label} & \textbf{Count} & \textbf{Percentage} \\
\hline
Text & 3,671 & 79.4\% \\
Page-footer & 663 & 14.2\% \\
Title & 229 & 4.9\% \\
Section-header & 59 & 1.2\% \\
\hline
\end{tabular}
\caption{Distribution of annotations for each class in the dataset}
\label{tab:annotation_distribution_02}
\end{table}

\subsection{Dataset for OCR}

In addition to annotations for layout analysis, we also created a dataset for OCR.
\subsubsection{Annotations for text line detection}

\begin{figure}[h!]
    \centering
    \includegraphics[width=1\linewidth]{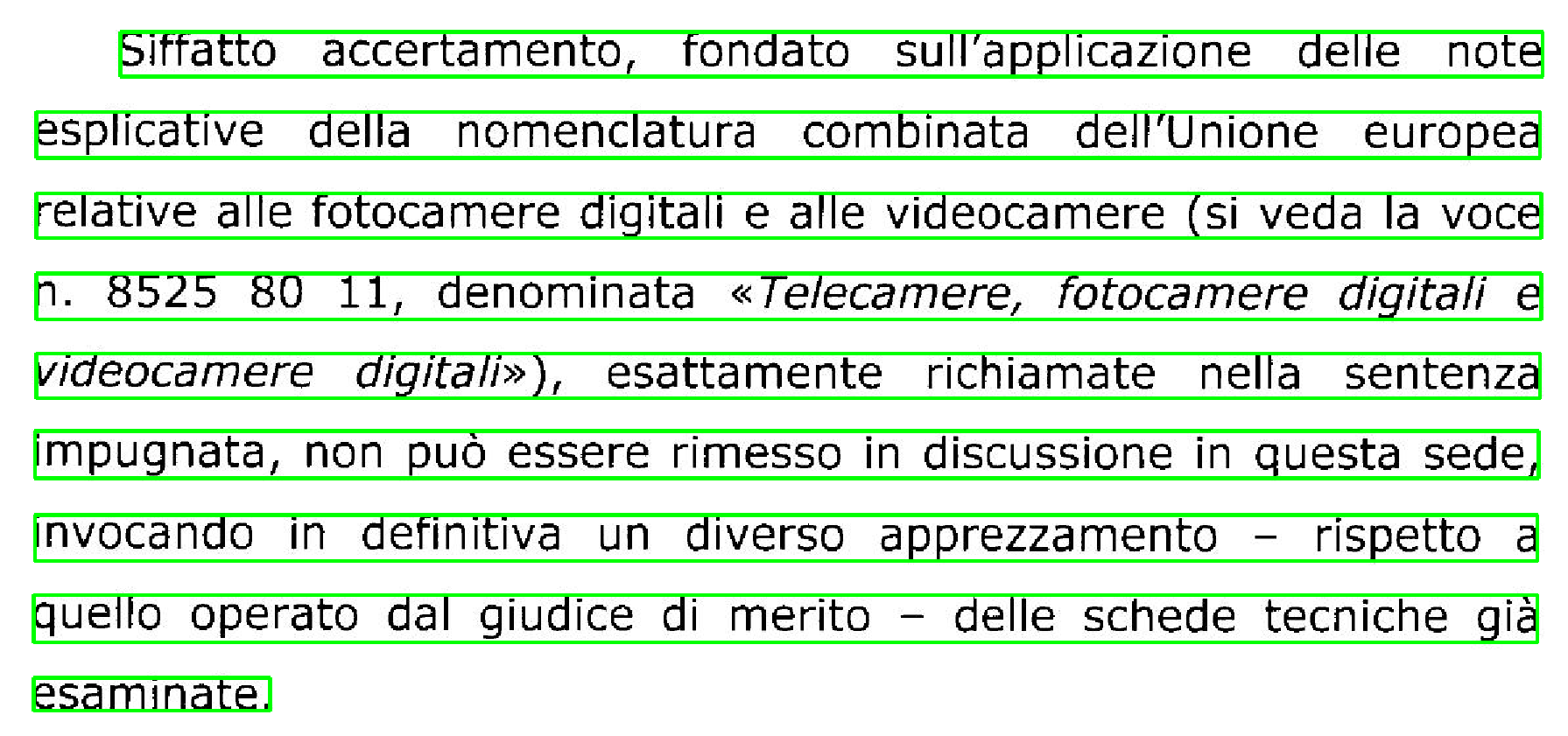}
    \caption{As an example, an annotation case for the text detection task is shown. The green bounding boxes represent ground truth and identify a row within the crop.}
    \label{fig:text-dect-ocr}
\end{figure}

To train the OCR text detection module, we extracted a crop dataset from paragraphs labeled as “Text” in the DLA dataset. Each crop was annotated with bounding boxes corresponding to text lines, represented by green bounding boxes, as illustrated in Figure \ref{fig:text-dect-ocr}. This dataset serves as the foundation for text line detection, ensuring precise localization of textual content. Table \ref{tab:ocr_text_line_dataset_distributions} provides an overview of the dataset size and details the distribution across different splits.

\begin{table}[h!]
\centering
\setlength{\tabcolsep}{6pt}
\renewcommand{\arraystretch}{1.2}
\begin{tabular}{l|ccc}
\hline
\textbf{Dataset Total} & \textbf{Training-set} & \textbf{Validation-set} & \textbf{Test-set} \\
\hline
10,442 & 8,341 (80\%) & 1,023 (10\%) & 1,078 (10\%) \\
\hline
\end{tabular}
\caption{Distribution of text line images in our OCR dataset, split according to an 80-10-10 ratio.}
\label{tab:ocr_text_line_dataset_distributions}
\end{table}

\subsubsection{Annotations for text recognition}

\begin{figure}[h!]
    \centering
    \includegraphics[width=1\linewidth]{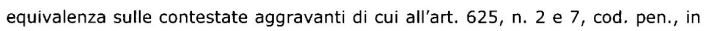}
    \caption{Line of text extracted from one of the sentences. }
    \label{fig:row-text-ocr}
\end{figure}

The text extraction module is trained on pairs of cropped images and their corresponding transcriptions. To construct this dataset, we generated line-level text crops and manually transcribed the content, as illustrated in Figure \ref{fig:row-text-ocr}. This process resulted in a total of 27,707 annotated samples extracted from the judgments. Table \ref{tab:ocr_dataset_distributions} provides a detailed breakdown of the dataset size across different splits.

\begin{table}[h!]
\centering
\setlength{\tabcolsep}{6pt}
\renewcommand{\arraystretch}{1.2}
\begin{tabular}{l|ccc}
\hline
\textbf{Dataset Total} & \textbf{Training-set} & \textbf{Validation-set} & \textbf{Test-set} \\
\hline
27,707 & 22,175 (80\%) & 2,754 (10\%) & 2,778 (10\%) \\
\hline
\end{tabular}
\caption{Distribution of text line images in our OCR dataset, split according to an 80-10-10 ratio.}
\label{tab:ocr_dataset_distributions}
\end{table}

\section{Methedology}
\subsection{Overview of the pipeline}
\begin{figure}
\centering
\includegraphics[width=1\linewidth]{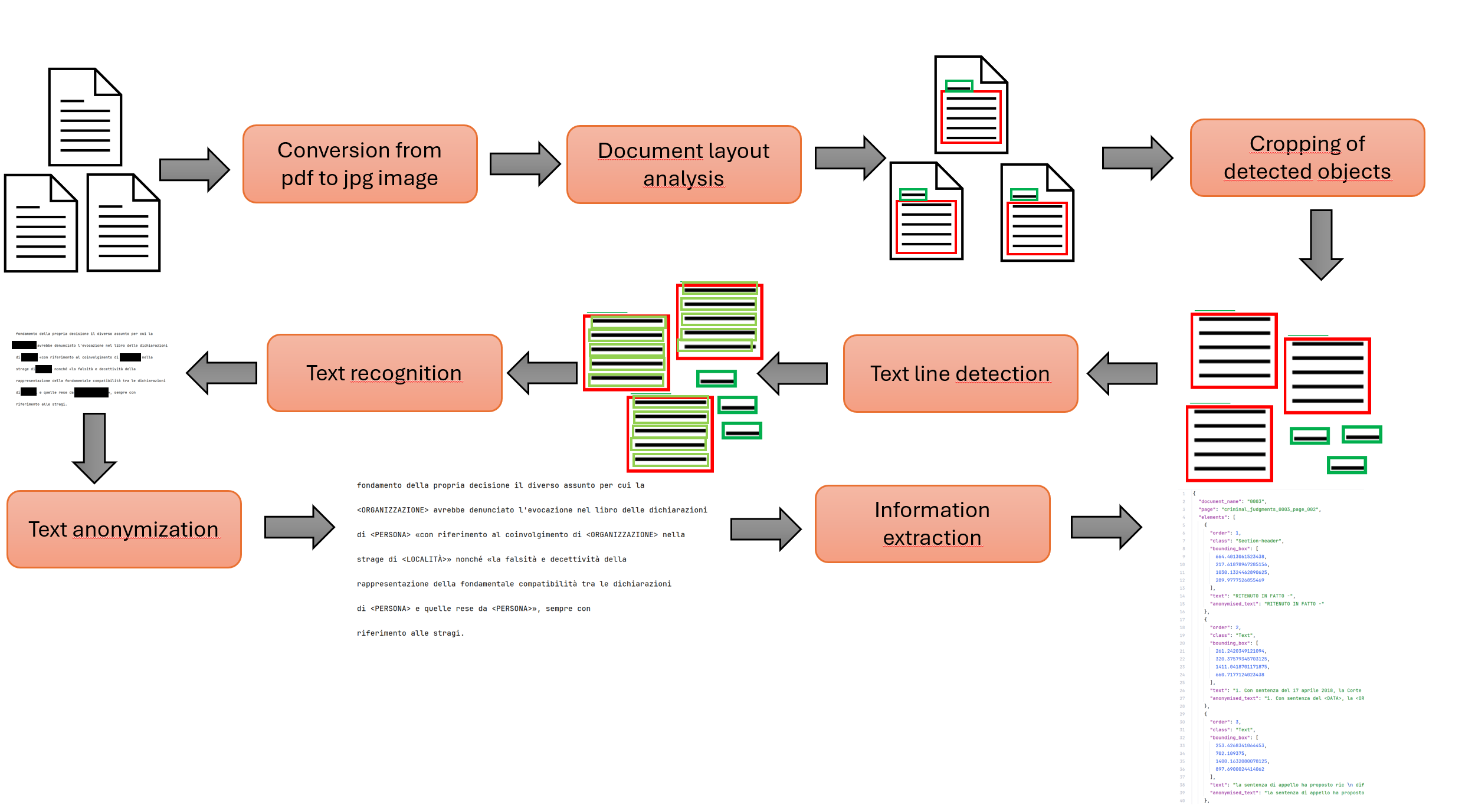}
\caption{Overview of our pipeline for document content analysis and extraction.}
\label{fig:pipeline_overview}
\end{figure}

Figure~\ref{fig:pipeline_overview} provides a comprehensive overview of our workflow for converting legal judgments from PDFs into anonymized textual data ready for further analysis. The pipeline consists of six sequential steps:

\begin{itemize}
    \item Preprocessing: converting PDFs into JPEG images.
    \item Document Layout Analysis: detecting structural components using YOLOv8 \cite{redmon2016you}.
    \item Text Line Detection: identifying individual lines of text with YOLOv8.
    \item Text Recognition: extracting machine-readable text using TrOCR \cite{li2023trocr}.
    \item Text Anonymization: masking sensitive information via GLiNER \cite{zaratiana2023gliner}.
    \item Information Extraction: storing document content in a structured, machine-readable format.
\end{itemize}

Each of these steps is crucial for ensuring high-quality text extraction and structured representation. In the following subsections, we provide a detailed discussion of the models, training procedures, and hyperparameters used at each stage.

\subsection{Document Preprocessing}
To facilitate layout analysis and text extraction, PDF documents were converted into JPEG images with a resolution of 200 DPI, ensuring optimal text readability. Each judgment was organized into a dedicated folder, with pages sequentially numbered to maintain a clear mapping to the original document. This structured organization preserves document integrity and simplifies downstream processing.

\subsection{Document Layout Analysis}
Document Layout Analysis was performed using the X version of YOLOv8 from Ultralytics. We selected YOLO due to its superior accuracy and efficiency, with version X ranking among one of most powerful detector models.

To leverage transfer learning, we initially trained YOLO on the DocLayNet dataset \cite{pfitzmann2022doclaynet}, aligning our annotation strategy with DocLayNet's labeling rules. This approach ensured compatibility and enabled more effective model adaptation. Given the limited size of our dataset, we mitigated overfitting by freezing the first 22 layers of YOLO during training.

The model was trained for 50 epochs, with early stopping set at 5 and a batch size of 8. We set the input resolution to 1024×1024 pixels, prioritizing high spatial fidelity to detect even small structural elements, such as page footers.

Table \ref{tab:yolo_hyper-parameters_for_dla_task} summarizes the hyperparameters used for fine-tuning. The trained model outputs bounding boxes, which are sorted to reflect the Western reading order. These ordered segments are then cropped and further processed for text extraction.

\begin{table}
\centering
\small
\setlength{\tabcolsep}{5pt}
\renewcommand{\arraystretch}{1.2}
\begin{tabular}{l|cccccc}
\hline
\textbf{Hyper-parameter} & \textbf{epoch} & \textbf{patience} & \textbf{batch-size} & \textbf{imgsz} & \textbf{freeze} & \textbf{worker} \\
\hline
\textbf{Value} & 50 & 5 & 8 & 1024 & 22 & 20 \\
\hline
\end{tabular}
\caption{Hyperparameters used for YOLOv8X fine-tuning on our dataset.}
\label{tab:yolo_hyper-parameters_for_dla_task}
\end{table}

The model outputs bounding boxes that delineate different structural components of the document such as: title, section-header, text, and page-footer. These bounding boxes are then sorted according to the Western reading order, ensuring a coherent sequence. Once ordered, the extracted regions undergo a cropping operation to facilitate accurate text line detection and text recognition.

\subsection{OCR}
The OCR module was designed by dividing it into two components:
\begin{itemize}
    \item Text line detection by YOLOv8.
    \item Text extraction by TrOCR.
\end{itemize}

\subsubsection{Text line detection}

Extracted document components were analyzed to identify individual lines of text. For this task, we employed YOLOv8 version X, initializing it with the pre-trained DocLayNet model. Given the ample size of our dataset, we opted to train all network layers, allowing the model to fully adapt to our domain. The hyperparameters remained unchanged from Table \ref{tab:yolo_hyper-parameters_for_dla_task}, except for the number of epochs, set to 25, and the number of trainable layers, as detailed in Table \ref{tab:yolo_hyper-parameters_for_textline_detec_task}.

\begin{table}
\centering
\small
\setlength{\tabcolsep}{5pt}
\renewcommand{\arraystretch}{1.2}
\begin{tabular}{l|cccccc}
\hline
\textbf{Hyper-parameter} & \textbf{epoch} & \textbf{patience} & \textbf{batch-size} & \textbf{imgsz} & \textbf{freeze} & \textbf{worker} \\
\hline
\textbf{Value} & 25 & 5 & 8 & 1024 & 0 & 20 \\
\hline
\end{tabular}
\caption{Hyperparameters used for YOLOv8X fine-tuning on text line dataset detection.}
\label{tab:yolo_hyper-parameters_for_textline_detec_task}
\end{table}

Once detected, the text lines were sorted according to the Western reading order and cropped to fit the text extraction module, ensuring a structured and coherent representation.
\subsubsection{Text extraction}
For the crucial task of converting detected text line into a raw text, we used TrOCR \cite{li2023trocr}, a powered transformer based text recognition model. Unlike traditional approaches that rely on CNN backbones and RNN-based decoders, TrOCR harnesses the power of pretrained Transformers for vision and language in an end to end architecture.

The model processes each clipped text line using an image transformer encoder, which segments the input into 16×16 pixel patches and extracts visual features. These features are then passed to a text transformer, which decodes them into word-level text.

For this task, we employed the TrOCR-Small configuration, which integrates BEiT \cite{bao2021beit} as the vision encoder and an auto-regressive decoder initialized with RoBERTa weights \cite{liu2019roberta}. This setup demonstrated superior performance in document text recognition, making it the optimal choice for our pipeline.

A key advantage of TrOCR in our use case is its ability to leverage context for accurate text generation. Unlike conventional OCR models, its pre-trained decoder incorporates strong language understanding, enabling it to exploit contextual cues while reconstructing text. This capability allows TrOCR to effectively handle variations in font styles and sizes, which are common in legal documents, while maintaining high recognition accuracy.

To further adapt the model to the Italian legal domain, we fine-tuned TrOCR-Small using Microsoft's pretrained version 'microsoft/trocr-small-printed' on our dataset. The hyperparameters used for fine-tuning are detailed in Table \ref{tab:trocr_hyper-parameters_for_text_extraction_task}.

\begin{table}[ht]
\centering
\setlength{\tabcolsep}{3pt}
\renewcommand{\arraystretch}{1.2}
\begin{tabular}{l|cccccc}
\hline
\textbf{Hyper-param.} & \textbf{epoch} & \textbf{batch} & \textbf{max-len} & \textbf{beams} & \textbf{len-pen.} & \textbf{no-repeat} \\
\hline
\textbf{Value} & 7 & 8 & 64 & 8 & 2 & 3 \\
\hline
\end{tabular}
\caption{Hyperparameters used for TrOCR fine-tuning on our dataset.}
\label{tab:trocr_hyper-parameters_for_text_extraction_task}
\end{table}

As stated in Subsection \ref{sec:Data_source}, the quality of the judgments received is satisfactory; however, printing defects, such as ink omissions, may sporadically occur.

\begin{figure}[!h]
    \centering
    \includegraphics[width=1\linewidth]{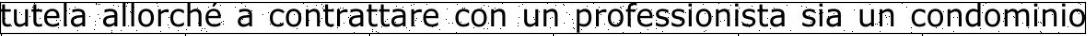}
    \caption{An example of our data augmentation}
    \label{fig:ocr_data_aug_ese_01}
\end{figure}

\begin{figure}[!h]
    \centering
    \includegraphics[width=1\linewidth]{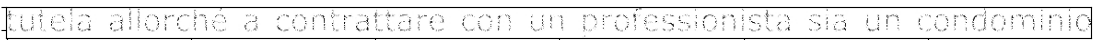}
    \caption{An example of our data augmentation}
    \label{fig:ocr_data_aug_ese_02}
\end{figure}

\begin{figure}[!h]
    \centering
    \includegraphics[width=1\linewidth]{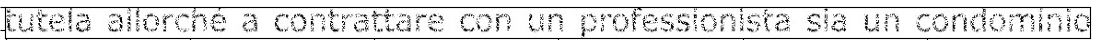}
    \caption{An example of our data augmentation}
    \label{fig:ocr_data_aug_ese_03}
\end{figure}

Consequently, we implemented a data augmentation technique to enhance TrOCR's generalization capabilities by simulating these ink defects. For illustrative purposes, Figures \ref{fig:ocr_data_aug_ese_01}, \ref{fig:ocr_data_aug_ese_02}, and \ref{fig:ocr_data_aug_ese_03} depict examples of this data augmentation.


\subsection{Text anonymization}
Although Supreme Court judgments are public and written without confidentiality constraints, ensuring the privacy and dignity of the subjects involved remains a priority. Anonymization not only protects sensitive information but also mitigates bias, preventing NLP models from memorizing overly specific patterns embedded in the data.

To achieve this, we focus on identifying and masking key entities, including parties, witnesses, companies, dates, and locations. These entities have been grouped into five categories, as detailed in Table \ref{tab:entity_table}.

\begin{table}[ht]
\centering
\setlength{\tabcolsep}{3pt}
\renewcommand{\arraystretch}{1.2}
\begin{tabular}{lcc}
\hline
\textbf{Entity} & \textbf{Tagging} & \textbf{Description}  \\
\hline
Organization & $\langle \text{ORGANIZAZZIONE}\rangle$ & Organizations, companies, agencies, institutions   \\
\hline
Person & $\langle \text{PERSONA}\rangle$ & Names of parties, lawyers, judges, family, experts, witness   \\
\hline
Location &   $\langle \text{LOCALIT\`A}\rangle$  & Place of birth, residence, addresses, countries, cities, states    \\
\hline
Email &  $\langle \text{EMAIL}\rangle$  & emails, PECs    \\
\hline
DATE &  $\langle \text{DATA}\rangle$  & Date of birth, marriage, death, date of events    \\
\hline
ID &  $\langle \text{ID}\rangle$  & Tax Codes, number plates, VAT numbers    \\
\hline
\end{tabular}
\caption{Entity name and description of the 5 entities to be detected}
\label{tab:entity_table}
\end{table}

For the crucial task of text anonymization, we rely on GLiNER, a Named Entity Recognition (NER) model designed to detect and classify entities in legal documents. GLiNER employs a BERT-like bidirectional transformer encoder, which processes text in both directions to capture rich contextual information.

The model generates embeddings for entity types, encodes text into a latent space, and aggregates tokens into span representations, enabling the recognition of multi-word entities. These representations are then evaluated by a classifier that identifies which text segments correspond to the specified entity types.

A key feature of GLiNER is its zero-shot classification capability, allowing it to recognize entity types never seen during training. This makes the model highly adaptable across different domains, including legal contexts with diverse and unpredictable entity forms.

In our work, we use an Italian-optimized version of GLiNER, fine-tuned to detect Personally Identifiable Information (PII), available from the Hugging Face Hub \cite{GLiNERita}. This configuration enables us to effectively identify and redact sensitive entities commonly found in legal documents.

\subsection{Information extraction}

After applying all processing steps, we store each page’s information in a JSON document, enabling seamless integration with downstream NLP applications, such as topic modeling. Each JSON file contains the page name and an "elements" list, where every document component is described by both spatial and semantic attributes. These include the bounding box coordinates in Pascal VOC format \footnote{In the Pascal VOC format, $x_{min}$ and $y_{min}$ are the coordinates of the top left corner of the bounding box, while $x_{max}$ and $y_{max}$ represent the coordinates of the bottom right corner.} $[x_{\min}, y_{\min}, x_{\max}, y_{\max}]$, the element class, the plain text, and the corresponding anonymized text.

\begin{figure}[htbp]
    \begin{minipage}{0.99\textwidth}
        \centering
        \includegraphics[width=\linewidth]{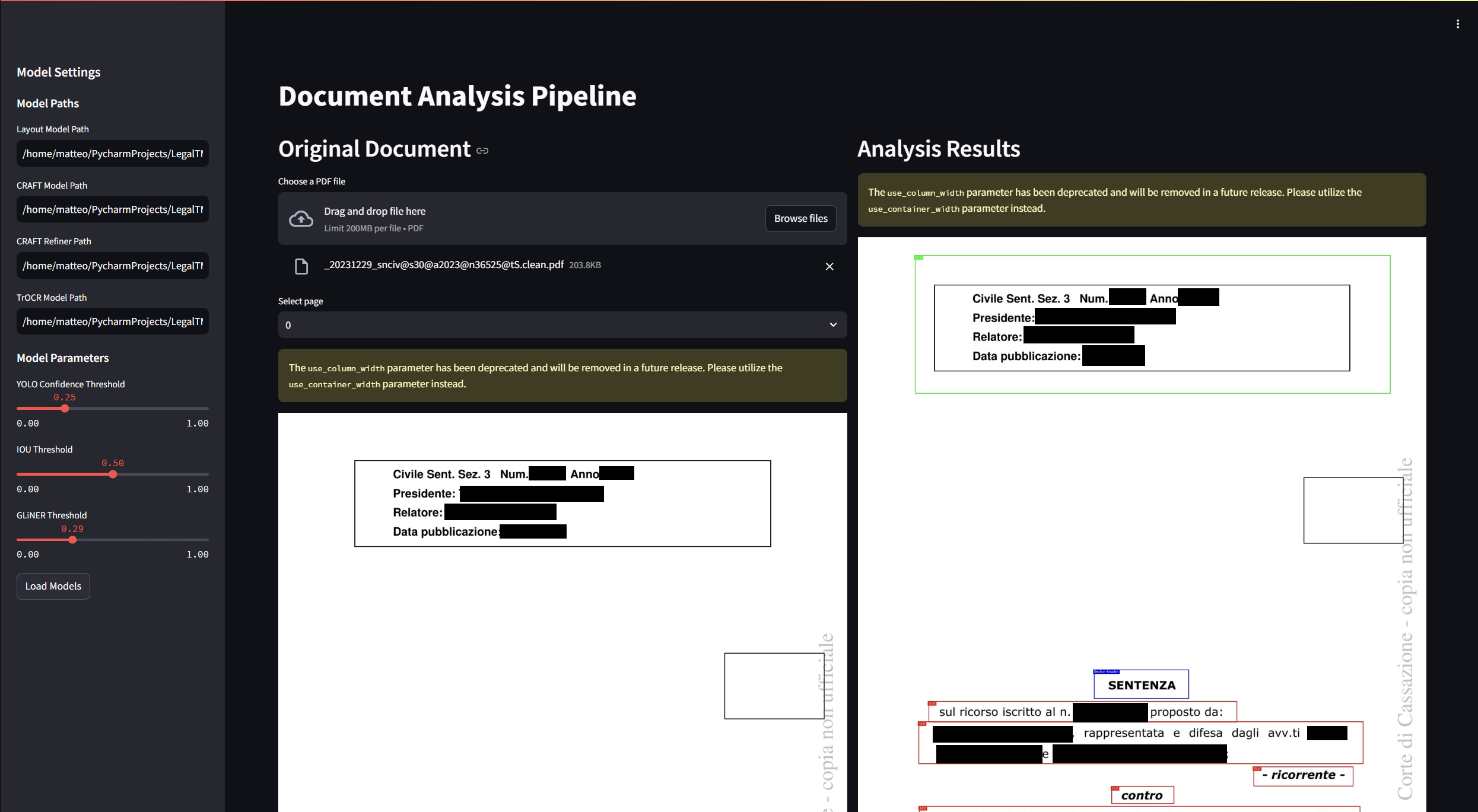}
        \caption{The streamlit app for our pipeline.}
        \label{fig:streamlit_app01}
    \end{minipage}
    \hfill
    \begin{minipage}{0.99\textwidth}
        \centering
        \includegraphics[width=\linewidth]{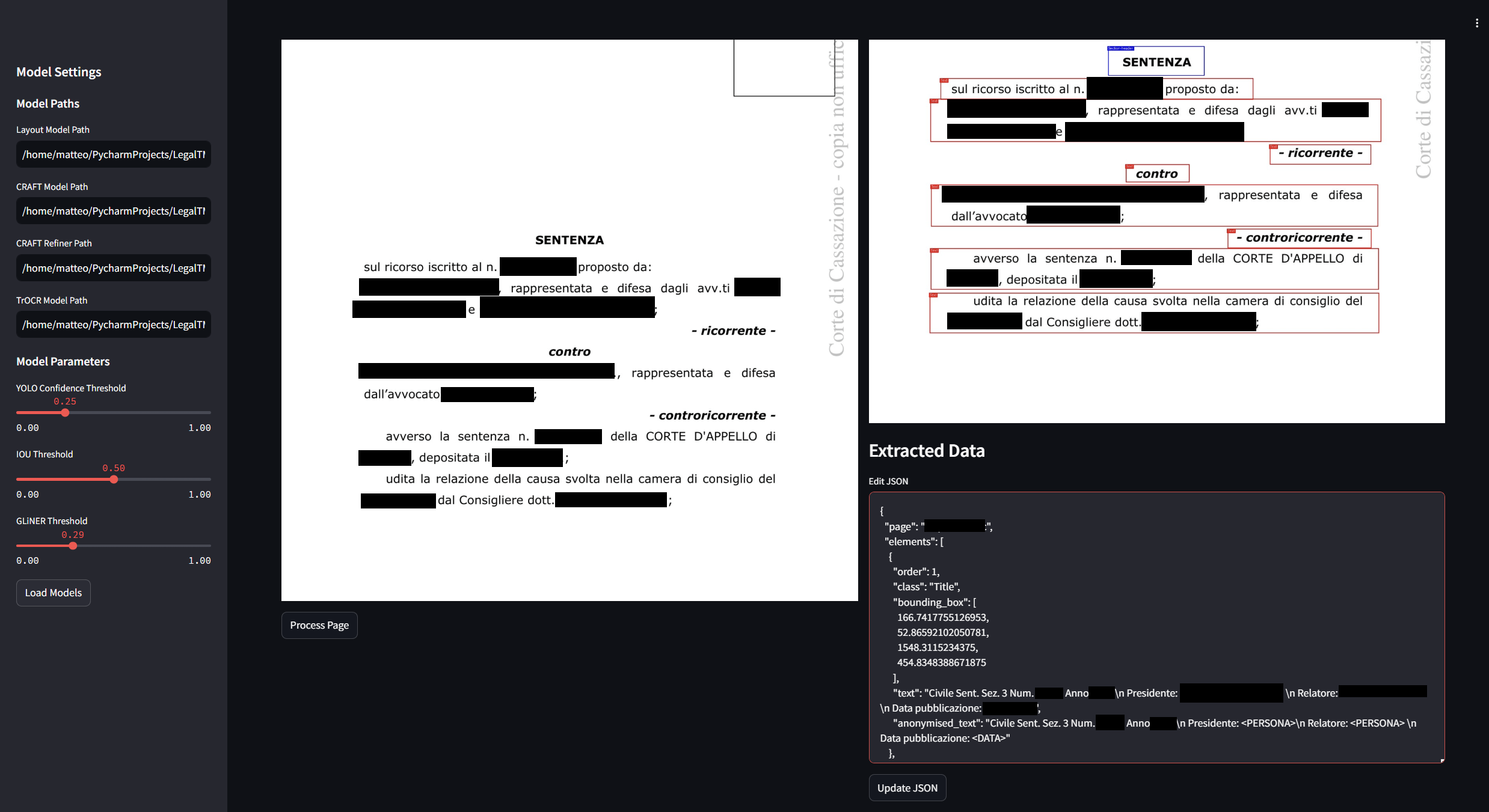}
        \caption{The streamlit app for our pipeline.}
        \label{fig:streamlit_app02}
    \end{minipage}
\end{figure}

To make the pipeline accessible to non-expert users, we developed an interactive Streamlit front end, shown in Figures \ref{fig:streamlit_app01} and \ref{fig:streamlit_app02}. The application allows users to load models and documents with ease, and to adjust key hyperparameters, such as YOLO’s confidence threshold and IoU, or GLiNER’s entity confidence threshold, to fine-tune model behavior.

Additionally, users can manually edit the JSON content, enabling them to review and correct the plain text or anonymized fields as needed, ensuring both flexibility and control over the final output.

\subsection{A novel topic modeling dataset}
After processing all available judgments, we compiled the final dataset by aggregating the generated JSON files.

\begin{figure}[h!]
    \centering
    \includegraphics[width=1\linewidth]{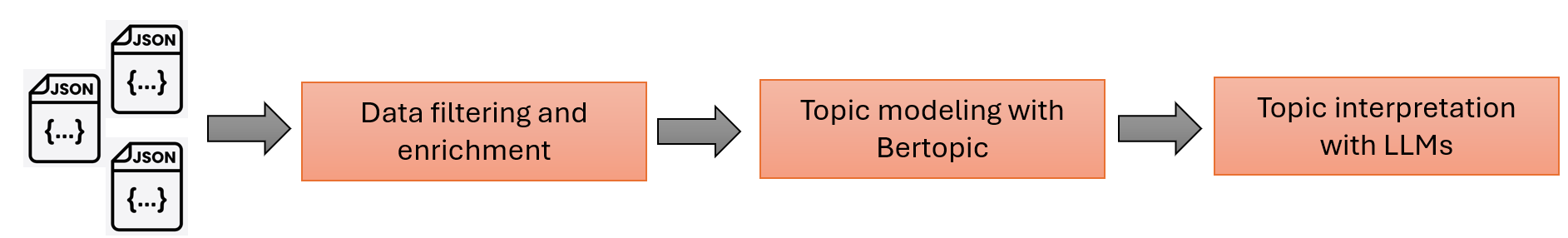}
    \caption{The pipeline for topic modeling.}
    \label{fig:topic_modeling_pipeline}
\end{figure}

As shown in Figure \ref{fig:topic_modeling_pipeline}, the topic modeling pipeline begins by extracting key elements from the JSON files, including the document name, page number, anonymized text, and content class. This information is then structured into a Pandas \cite{mckinney2011pandas} DataFrame, enabling further enrichment with additional metadata, such as text length.

To refine the dataset for analysis, we filtered out text belonging to the Title, Section-header, and Page-footer classes, as these elements were not relevant to our study.

\subsection{Topic modeling}
Topic modeling was performed using BERTopic \cite{grootendorst2022bertopic}, chosen for its high modularity, which enables customization by replacing or adding components based on task-specific requirements.

To generate text embeddings, we replaced the standard transformer with Distilled Legal-Italian BERT \cite{licari2024italian}, a model specialized for the Italian legal domain. This choice was driven by the linguistic characteristics of our dataset, ensuring better representation of legal terminology.

The resulting embeddings, typically high-dimensional (768 dimensions) and dense, were then processed with UMAP \cite{ghojogh2021uniform}, a dimensionality reduction technique. While measures like cosine distance can help mitigate high-dimensionality issues, clustering algorithms still struggle to extract meaningful structures from such large feature spaces. Reducing dimensionality improves computational efficiency and enhances cluster separability.

Finally, we applied HDBSCAN \cite{mcinnes2017hdbscan} to cluster the reduced embeddings. As an advanced variant of DBSCAN, HDBSCAN offers greater speed and robustness, particularly in datasets with variable-density clusters, making it well-suited for text data.

BERTopic identifies the characteristic words of each cluster through a structured three-step process:
\begin{enumerate}
    \item Stop word removal: Eliminating frequent, non-informative terms such as “courts,” “lawyers,” “firms,” “judgments,” and anonymized words.
    \item Bag-of-Words (BoW) model construction \cite{qader2019overview}, representing text as token frequency distributions.
    \item c-TF-IDF computation \cite{grootendorst2022bertopic}, a cluster-level adaptation of TF-IDF that enhances word importance estimation.
\end{enumerate}

In particular, we employed the c-TF-IDF-bm25 variant, which is optimized for small datasets and offers improved handling of stop words, ensuring better topic representation.

The hyperparameters used in the BERTopic pipeline are listed in Table \ref{tab:hyperparameters}.

\begin{table}[htbp]
\centering
\setlength{\tabcolsep}{8pt}
\renewcommand{\arraystretch}{1.2}
\begin{tabular}{l|l}
\hline
\textbf{Algorithm} & \textbf{Hyper-parameters} \\
\hline
\multirow{3}{*}{Embedding Model} & id\_model = dlicari/distil-ita-legal-bert \\
                                & max\_seq\_length = 512 \\
                                & batch\_size = 32 \\
\hline
\multirow{4}{*}{UMAP} & n\_components = 5 \\
                    & min\_dist = 0.0 \\
                    & metric = cosine \\
                    & n\_neighbors = 5 \\
\hline
Count Vectorizer    & ngram\_range = (1, 2) \\
                    & stop\_words = stopwords \\
\hline
\multirow{3}{*}{HDBSCAN} & min\_cluster\_size = 5 \\
                        & metric = euclidean \\
                        & min\_samples = 5 \\
\hline
\multirow{2}{*}{Topic Extraction} & top\_n\_words = 15 \\
                                 & diversity = 0.35 \\
\hline
\end{tabular}
\caption{Configuration of hyperparameters for the models that make up BERTopic.}
\label{tab:hyperparameters}
\end{table}

\subsection{Interpretation of topics with LLMs}
Below, we present the interpretation of topics obtained with BERTopic, enhanced by the use of Large Language Models (LLMs). Since OpenAI introduced the widely popular ChatGPT, interest in Generative Artificial Intelligence has surged across academia, industry, and the general public. This growing attention is driven by the remarkable capabilities of these models in both interpreting and generating content in response to user-defined prompts.

Integrating LLMs with topic modeling techniques such as BERTopic enables a deeper analysis of extracted topics, providing contextualized interpretations of patterns within legal documents. In addition to these powerful LLMs, we also use LLaMA 3.1 8B \cite{dubey2024llama} in this research. LLaMA 3.1 8B can run nimbly locally on machines with modest hardware and can understand and generate Italian text.

Given that our data were anonymized, we also integrated powerful closed-source LLMs, such as Claude 3.7 Sonnet \cite{anthropicclaude3} and GPT-4 \cite{achiam2023gpt}. In these cases, anonymization is crucial, as these models operate on private servers located outside Italy and the European Union.

To conduct our analysis, we designed two prompts in Italian. The first prompt generates topic labels by analyzing the most representative keywords and a set of sample documents:

\begin{quote}
\textit{""" Sei un esperto di giurisprudenza e di diritto italiano.
Rispondi in modo professionale alle richieste.
Ho un topic descritto dalle seguenti keywords: [KEYWORDS]
In questo topic, i seguenti documenti sono un sottoinsieme piccolo ma rappresentativo di tutti i documenti dell'argomento: [REPR\_DOCS].
Sulla base delle informazioni di cui sopra, fornisci una breve label a questo topic.
Assicurati di riportare solo la label e nient'altro. """}
\end{quote}

The second prompt is designed to generate a brief descriptive summary of each topic:

\begin{quote}
\textit{""" Sei un esperto di giurisprudenza e di diritto italiano.
Rispondi in modo professionale alle richieste.
Ho un topic descritto dalle seguenti keywords: [KEYWORDS]
In questo topic, i seguenti documenti sono un sottoinsieme piccolo ma rappresentativo di tutti i documenti dell'argomento: [REPR\_DOCS].
Sulla base delle informazioni di cui sopra, fornisci una descrizione di questo topic nel seguente formato:
topic: $<descrizione>$ """}
\end{quote}

Within BERTopic, keywords refer to the terms automatically extracted by the model to define each topic.
Conversely, Representative Documents correspond to exemplar texts that best encapsulate the core characteristics of each topic.

For text generation, Claude-Sonnet 3.7 and GPT-4 were used with default hyperparameters, while LLaMA 3.1 was fine-tuned with the configurations listed in Table \ref{tab:llama_hyperparams}, both for label generation and text summarization tasks.

\begin{table}[h]
\centering
\setlength{\tabcolsep}{8pt}
\renewcommand{\arraystretch}{1.2}
\begin{tabular}{l|ccc}
\hline
\textbf{Task} & \textbf{max\_new\_tokens} & \textbf{temperature} & \textbf{repetition\_penalty} \\
\hline
Label Generation & 50 & 0.1 & 1.1 \\
Text Summary & 2048 & 0.1 & 1.1 \\
\hline
\end{tabular}
\caption{Hyperparameters for Llama 3.1 (8B) model tasks}
\label{tab:llama_hyperparams}
\end{table}

To conclude this subsection and the entire section, we present the pseudocode of our topic modeling algorithm in Algorithm \ref{alg:topic_modeling}.

\begin{figure*}[!th]
    \centering
    \begin{minipage}{0.9\textwidth}
        \begin{algorithm}[H]
            \caption{Topic Modeling Algorithm}
            \begin{algorithmic}[1]
                \State \textbf{Input:} 
                \begin{itemize}
                    \item Let \(D = \left\{ \{p_{11}, p_{12}, \dots, p_{1m_1}\}, \dots, \{p_{n1}, p_{n2}, \dots, p_{nm_n}\} \right\}\) a set of anonymized textual data
                \end{itemize}
                \State \textbf{Output:} 
                \begin{itemize}
                    \item Identified topics with labels and description
                \end{itemize}
                \State \textbf{Steps:}
                \begin{enumerate}
                    \item \textbf{Embedding Generation with Distiled Italian-legal BERT}
                    \begin{itemize}
                        \item \(\mathbf{e}_{ij} = M(p_{ij}) \quad \forall i = 1, 2, \dots, n \quad \wedge \quad \forall j = 1, 2, \dots, m_i\)
                    \end{itemize}
                    \item \textbf{Dimensionality Reduction with UMAP}
                    \begin{itemize}
                         \item \( E = \left\{ \{\mathbf{e}_{11}, \mathbf{e}_{12}, \dots, \mathbf{e}_{1m_1}\}, \{\mathbf{e}_{21}, \mathbf{e}_{22}, \dots, \mathbf{e}_{2m_2}\}, \dots, \{\mathbf{e}_{n1}, \mathbf{e}_{n2}, \dots, \mathbf{e}_{nm_n}\} \right\} \)
                        \item \(\mathbf{z}_{ij} = \text{UMAP}(\mathbf{e}_{ij}) \quad \forall i = 1, 2, \dots, n \quad \wedge \quad  \forall j = 1, 2, \dots, m_i\)
                    \end{itemize}
                    \item \textbf{Clustering embeddings with HDBSCAN}
                    \begin{itemize}
                        \item \( Z = \left\{ \{\mathbf{z}_{11}, \mathbf{z}_{12}, \dots, \mathbf{z}_{1m_1}\}, \{\mathbf{z}_{21}, \mathbf{z}_{22}, \dots, \mathbf{z}_{2m_2}\}, \dots, \{\mathbf{z}_{n1}, \mathbf{z}_{n2}, \dots, \mathbf{z}_{nm_n}\} \right\} \)
                        \item $ C = \text{HDBSCAN}(Z) $
                        \item $ C = \{ C_1, C_2, \dots, C_k, -1 \}$
                    \end{itemize}
                    \item \textbf{Topic Representation with using c-TF-IDF}
                    \begin{itemize}
                        \item $w_{x, C_i} = \text{tf}_{x, C_i} \times \log\left( 1 + \frac{A- f_x + .5}{f_x + .5} \right)$
                        \item $w_{x, C_i}$ represents the weight of the term $x$ in cluster $C_i$,
                        \item $\text{tf}_{x, C_i}$ is the frequency of the term $x$ within cluster $C_i$,
                        \item $f_x$ is the frequency of the term $x$ across all clusters,
                        \item $A$ is the average number of words per cluster (documents).
                    \end{itemize}
                    
                    \item \textbf{Generate labels using a LLMs }
                    \item \textbf{Generate text-summarization using a LLMs}
                \end{enumerate}
            \end{algorithmic}
            \label{alg:topic_modeling}
        \end{algorithm}
    \end{minipage}
\end{figure*}

\section{Evalutation}
\subsection{Metrics}
In this section we report metrics to evaluate our pipeline, topic modeling metrics, and text generation metrics for LLMs. 
\subsubsection{Object detection metrics}
Object detection performance is typically assessed using a combination of metrics that evaluate both localization quality (bounding box) and classification accuracy of the detected instances.

To evaluate the accuracy of our DLA model, we primarily use mean Average Precision (mAP), computed at a 50\% Intersection over Union (IoU) \footnote{Given two rectangles, R1 (representing the ground truth) and R2 (the model's prediction), Intersection over Union (IoU) measures the area of intersection between R1 and R2. If the two rectangles do not overlap, the IoU is 0; if they overlap perfectly, the IoU is 1.} threshold (mAP50). Additionally, we assess performance across multiple IoU thresholds ranging from 50\% to 95\% (mAP50-95), following the standard procedure in \cite{9145130}.

\paragraph{Precision and Recall}
Before describing mAP, it is helpful to introduce \emph{precision} and \emph{recall}, which measure how accurate and complete the model’s predictions are:
\begin{itemize}
    \item Precision is the fraction of predictions that are correct (i.e., true positives\footnote{True positives are model predictions that correspond to the truth class of the detected object and satisfy the IoU threshold.}) among all predicted instances (the sum of true positives and false positives \footnote{False positives are detections that do not correspond to the class of the detected object or that do not exceed the IoU threshold.}). Formally:
    \begin{equation}
        Precision = \frac{\text{True Positives}}{\text{True Positives} + \text{False Positives}}.
    \end{equation}
    \item Recall  is the fraction of objects that are correctly identified (true positives) among all instances that should have been detected (the sum of true positives and false negatives \footnote{False negative are detections that are not taken}). Formally:
    \begin{equation}
        Recall = \frac{\text{True Positives}}{\text{True Positives} + \text{False Negatives}}.
    \end{equation}
\end{itemize}

Both metrics are crucial in object detection: high precision indicates a low number of spurious detections (i.e., few false positives), while high recall means the model misses few objects (i.e., few false negatives).

In object detection tasks, a prediction is typically considered a true positive if the IoU between the predicted bounding box and the ground-truth bounding box exceeds a given threshold (e.g., 50\%).

\paragraph{Average Precision (AP)}
To combine precision and recall into a single metric, Average Precision (AP) is used. It is computed as the area under the precision-recall curve for a single class.

An AP value close to 1 indicates that, across various model confidence thresholds, precision remains high over a wide range of recall values. Typically, AP is computed for each class and then averaged across all classes of interest, yielding the mean Average Precision (mAP)."

\paragraph{mAP50 and mAP50-95}
The calculation of mAP depends on the IoU threshold(s) used to consider a detection correct:
\begin{itemize}
    \item mAP50 is computed by evaluating the Average Precision for each class at a single IoU threshold (commonly 0.5) and then averaging these values. This metric is less stringent, as an IoU of at least 0.5 is enough to count a detection as correct.
    \item mAP50-95 is computed by averaging the AP values over multiple IoU thresholds (usually from 0.5 to 0.95 in increments of 0.05). This is more demanding, as it requires more precise overlap between the predicted bounding box and the ground-truth object.
\end{itemize}

Formally, mAP can be expressed as:
\begin{equation}
    \text{mAP} = \frac{1}{K} \sum_{i=1}^{K} \mathrm{AP}_i,
    \label{eq:mAP50}
\end{equation}
where $K$ is the number of classes and $\mathrm{AP}_i$ is the Average Precision for the $i$-th class.

\subsubsection{OCR metrics for text recognition task}
For the text recognition model, we employ two metrics \cite{neudecker2021survey} commonly used in OCR: Character Error Rate (CER) and Word Error Rate (WER).

\paragraph{Character Error Rate (CER)}
Character Error Rate (CER) is defined as the ratio of the total number of character-level errors (substitutions, deletions, and insertions) to the total number of characters in the ground truth:

\begin{equation}
        \text{CER} = \frac{\text{S} + \text{D} + \text{I}}{N}
\end{equation}

where \( \text{S} \) is the number of substitutions, \( \text{D} \) is the number of deletions, \( \text{I} \) is the number of insertions, and \( N \) is the total number of characters in the reference text. A lower CER indicates better accuracy at the character level.

\paragraph{Word Error Rate (WER)}
Word Error Rate (WER) measures the proportion of word-level errors in the recognized text:
\begin{equation}
    \text{WER} = \frac{\text{S} + \text{D} + \text{I}}{M}
\end{equation}

where \( \text{S} \), \( \text{D} \), and \( \text{I} \) are the number of substitutions, deletions, and insertions at the word level, and \( M \) is the total number of words in the reference text. A lower WER indicates better accuracy at the word level.

These metrics allow us to quantify the performance of the OCR module at both character and word level, and for both metrics, values close to 0 indicate that the text recognition model makes few errors and is reliable.

\subsubsection{Topic modeling metrics}
We evaluate the quality of the topics generated by our topic model using topic coherence and topic diversity.

\paragraph{Topic coherence $C_{v}$}

Topic coherence measures the degree of semantic similarity between high-probability words in a topic. Among the various coherence measures, we employ the $C_v$ metric \cite{roder2015exploring}, which is known to correlate well with human interpretability of topics. The $C_v$ coherence score is based on the normalized pointwise mutual information (NPMI) between pairs of top words in a topic.

The NPMI score between two words $x_{i}$ and $x_{j}$ is defined as follows:

\begin{equation}
    \text{NPMI}(x_i, x_j) = \frac{\log \frac{p(x_i, x_j) + \epsilon}{p(x_i)p(x_j)}}{-\log p(x_i, x_j) + \epsilon}
\end{equation}

where $p(x_{i}, x_{j})$ represents the co-occurrence probability of words $x_{i}$ and $x_{j}$ in a reference corpus, and $\epsilon$ is a smoothing constant used to avoid division by zero.

The coherence score for a topic $t_{k}$ is then calculated by averaging the NPMI values for all word pairs within the top words of the topic.

To obtain a single coherence score for the entire model, the $C_{v}$ score is computed as the mean coherence score across all topics K:

\begin{equation}
    C_v = \frac{1}{T} \sum_{i=1}^{T} \cos(\mathbf{v}_{\text{NPMI}}(x_i), \mathbf{v}_{\text{NPMI}}(x_j))
\end{equation}

where $V_{NPMI}$ represents the NPMI score vectors for each topic word.

A $C_{v}$ score close to 1 indicates that the keywords effectively describe the topics and are semantically related. Conversely, a value near 0 indicates that the keywords do not accurately represent the topics and are not semantically related.

\paragraph{Topic diversity}

Topic diversity \cite{dieng-etal-2020-topic} is a metric that quantifies how semantically diverse the obtained topics are. It is defined as follows:
\begin{equation}
TD = \frac{\left| \bigcup_{k} T_j \right|}{K \times N}
\label{eq:topic_diversity}
\end{equation}

where \( T_j \) represents the set of the first  words in topic \( j \), \( K \) is the total number of topics, and \( N \) is the number of words considered for each topic. The numerator represents the number of unique words across all topics.

The topic diversity metric ranges from 0 to 1, where 0 indicates an exact match between all topics in terms of their primary lexical elements, while 1 denotes complete divergence in vocabulary across all topics.

A topic diversity value close to 1 suggests that the keywords describing the topics are highly diverse, meaning that none or very few topics share keywords. Conversely, a value close to 0 indicates that many topics share a significant number of keywords.

\paragraph{Trade-off between topic diversity and topic coherence}

Topic diversity and topic coherence have an inverse relationship \cite{bilal2021evaluation}, \cite{wu-etal-2020-short}. Maximizing one of these metrics typically results in minimizing the other: models that maximize topic coherence tend to generate topics with highly similar main words, leading to reduced keyword diversity. As a result, topics become too similar.

Conversely, a model that maximizes topic diversity will produce topics that are very different from each other, incorporating a broader range of keywords but potentially reducing the interpretability of the generated topics. Therefore, it is essential to find a balance between the two metrics."

\subsubsection{Text generation metrics}
To evaluate labels and summaries for topics, we chose the BERTScore metric \cite{zhang2019bertscore} for LLMs. Unlike traditional metrics based on n-gram counts, such as ROUGE or BLEU, BERTScore can capture deeper semantic similarities between texts.

By leveraging BERT's contextual embeddings, this metric can recognize synonyms, paraphrases, and reformulations that preserve meaning while using different words. Additionally, BERTScore has shown a stronger correlation with human judgments than classical metrics, making it particularly suitable for evaluating generative outputs such as summaries and labels, where expressive flexibility is as important as content accuracy.

BERTScore computes the similarity between a generated text and a reference text using contextualized embeddings. Given a generated sentence \( X = \{x_1, x_2, \dots, x_m\} \) and a ground truth sentence \( Y = \{y_1, y_2, \dots, y_n\} \), their corresponding embeddings are obtained using a pre-trained transformer model:

\begin{equation}
    \mathbf{x}_i = \text{BERT}(x_i), \quad \mathbf{y}_j = \text{BERT}(y_j)
\end{equation}

where \( \mathbf{x}_i \) and \( \mathbf{y}_j \) are the embedding vectors for tokens \( x_i \) and \( y_j \), respectively.

BERTScore measures the pairwise cosine similarity between token embeddings:

\begin{equation}
    s_{ij} = \frac{\mathbf{x}_i \cdot \mathbf{y}_j}{\|\mathbf{x}_i\| \|\mathbf{y}_j\|}
\end{equation}

where \( s_{ij} \) represents the cosine similarity between the embedding of the \( i \)-th token in the generated text and the \( j \)-th token in the reference text.

BERTScore aggregates these similarity scores using precision, recall, and F1-score.

\paragraph{BERTScore: Precision}
BERTscore's Precision measures how alligned the sentence generated by the model with the ground truth sentence.

\begin{equation}
    P = \frac{1}{m} \sum_{i=1}^{m} \max_{j} s_{ij}
\end{equation}

A value close to 1 indicates that the generated sentence is able to capture the meaning of the ground truth sentence with high accuracy. While a value close to 0 indicates a significant semantic distance between the generated sentence and the ground truth sentence.

\paragraph{BERTScore: Recall}
BERTscore's recall measures how much of the meaning of the ground truth sentence is present in the sentence generated by the model.

\begin{equation}
    R = \frac{1}{n} \sum_{j=1}^{n} \max_{i} s_{ij}
\end{equation}

A value close to 1 indicates that almost all the information in the ground truth sentence has been captured in the generated sentence. While a recall value close to 0 indicates the generated sentence lacks much content present in the ground truth sentence.

\paragraph{BERTScore: F1-score}
BERTscore's F1-score is the harmonic mean of precision and recall, often used as a final score.

\begin{equation}
    F_1 = 2 \times \frac{P \times R}{P + R}
\end{equation}

A value close to 1 indicates that the generated sentence contains all the essential information present in the ground truth sentence and is very close in wording to that of ground truth, without adding irrelevant information.
While a value close to 0 indicates that the generated sentence does not contain all the essential information present in the ground truth sentence and is not very close in wording to the ground truth sentence, without adding irrelevant information.

\subsection{Hardware requirements}
The following results were obtained with a computer that has the following hardware configuration:
\begin{itemize}
    \item Processor: 13th Gen Intel(R) Core(TM) i9-13900H 2.60 GHz;
    \item GPU: NVIDIA RTX 4090 mobile, 16GB GDDR6;
    \item RAM: 32.0 GB
\end{itemize}

\subsection{Results}

Our comprehensive evaluation, conducted across multiple performance metrics, offers valuable insights into the effectiveness of each component within the pipeline. The results highlight both the strengths of individual modules and the areas that require further improvement, guiding future refinements of the system.

\subsubsection{Object detection evalutations}
Evaluation on the test set confirms the strong performance of our document layout analysis model across all content classes. The model consistently detects all structural components of the documents with a high average recall of 0.960, indicating its ability to identify nearly all relevant elements. Precision is also strong, reaching 0.933, as incorrect classifications are rare. The predicted bounding boxes align closely with the actual layout components, as reflected by an overall mAP@50–95 of 0.800, underscoring the model’s accuracy in localization.

A class-wise analysis reveals particularly noteworthy results. The “Title” class achieves a perfect recall of 1.000, meaning all titles are correctly detected, paired with a precision of 0.902 and the highest mAP@50–95 score of 0.959 among all classes. These metrics highlight the model’s exceptional ability to both recognize and precisely localize document titles.

In the case of the “Text” class, which dominates the dataset (411 out of 509 instances), the model delivers balanced and robust performance, with a precision of 0.954, recall of 0.951, and mAP@50–95 of 0.892. This confirms the model’s reliability in identifying the main textual content of legal documents.

The “Page-footer” class shows high precision (0.954), but slightly lower recall (0.926), indicating that footers are occasionally missed. Interestingly, while the mAP@50 is very high (0.982), the mAP@50-95 drops to 0.612, suggesting that the general location of footers is well predicted, but the precision of the selection rectangle can be improved. This result on the “Page-footer” class suggests that our model has problems with document objects being small. Fortunately, the “Page-footer” text of the judgments does not contain important information.

For the “Section-header” class, both precision (0.923) and recall (0.960) are in line with the global average. However, the mAP@50–95 score of 0.738 indicates that there is still room for improvement in precisely localizing these elements.

Overall, these results confirm the effectiveness and robustness of our layout analysis model, Table \ref{tab:table_dla_results_on_test_set}. It delivers consistently high performance across all structural classes, with particularly strong results for titles and main text, which are crucial for downstream NLP tasks.

\begin{table}
\centering
\setlength{\tabcolsep}{5pt}
\renewcommand{\arraystretch}{1.2}
\begin{tabular}{l|cc|cccc}
\hline
\textbf{Class} & \textbf{Images} & \textbf{Instances} & \textbf{Precision} $\uparrow$ & \textbf{Recall} $\uparrow$ & \textbf{mAP50} $\uparrow$ & \textbf{mAP50-95} $\uparrow$ \\
\hline
All & 82 & 509 & 0.933 & 0.960 & 0.964 & 0.800 \\ 
Page-footer & 67 & 67 & 0.954 & 0.926 & 0.982 & 0.612 \\ 
Section-header & 24 & 25 & 0.923 & 0.960 & 0.905 & 0.738 \\ 
Text & 81 & 411 & 0.954 & 0.951 & 0.975 & 0.892 \\ 
Title & 6 & 6 & 0.902 & 1.000 & 0.995 & 0.959 \\ 
\hline
\end{tabular}
\caption{The table shows the results of the document layout analysis model developed to examine the layout of judgments on test-set. Having treated the layout analysis problem as an object detection problem, evaluation metrics from that domain are used to evaluate the model. The arrow indicates how the metrics should be read. The higher the better.}
\label{tab:table_dla_results_on_test_set}
\end{table}

\subsubsection{OCR evalutations}

The OCR module was developed by integrating two key components: text detection and text recognition. To assess the system’s overall effectiveness, we evaluated each component separately, focusing on their individual contributions to the OCR task.

The text detection model, based on YOLOv8x, demonstrates exceptional performance, achieving a precision of 0.9864 and a recall of 0.9834. These results indicate that the model accurately identifies nearly all textual content in the documents, while minimizing false positives. The quality of localization is confirmed by a mAP@50 of 0.9901 and a mAP@50–95 of 0.9022, reflecting the precision of the predicted bounding boxes.

The text recognition module, implemented with TrOCR-Small, also achieves high transcription accuracy. The model records a Character Error Rate (CER) of 0.0047, showing that character-level recognition errors are minimal. Likewise, the Word Error Rate (WER) of 0.0248 confirms that the vast majority of words are correctly transcribed.

The combination of these two modules results in a robust and reliable OCR system, capable of both accurate localization and high-fidelity transcription of legal text. Performance metrics for the OCR pipeline are reported in Table \ref{tab:ocr_test_set_result_combined}.

Table \ref{tab:ocr_test_set_result_combined} presents the results on the test set for both the text detection and text recognition tasks.

\begin{table}[h]
\centering
\setlength{\tabcolsep}{2pt}
\renewcommand{\arraystretch}{1.2}
\begin{tabular}{l|c|cccc|cc}
\hline
\textbf{Model} & \textbf{Task} & \textbf{Precision} $\uparrow$ & \textbf{Recall} & \textbf{mAP50} $\uparrow$ & \textbf{mAP@50-95} $\uparrow$ & \textbf{CER} $\downarrow$ & \textbf{WER} $\downarrow$ \\ 
\hline
YOLOV8x & Text detection & 0.9864 & 0.9834 & 0.9901 & 0.9022 & - & - \\ 
TrOCR-Small & Text recognition & - & - & - & - & 0.0047 & 0.0248 \\ 
\hline
\end{tabular}
\caption{The table reports the results of the OCR model developed to detect and and extract the text of sentences on test-set. It consists of two rows in the table. In the first row are the results related to the text detection module, and in the second row are the results of the text recognition module. The direction of the arrow helps in the interpretation of the metrics. }
\label{tab:ocr_test_set_result_combined}
\end{table}

\subsubsection{The impact of the pipeline on Topic modeling}
In this sub section, we analyze the impact of the pipeline on the performance of BERTopic in topic modeling. To this end, we prepared two versions of the dataset for comparison. The first consists of documents processed exclusively with the OCR module, without further refinement. The second was generated using our full pipeline, then filtered to remove nonessential content—such as titles, page footers, section headers, and short or uninformative paragraphs, which often contain run-on or fragmented text.

\begin{figure}[!h]
    \centering
    \includegraphics[width=0.99\linewidth]{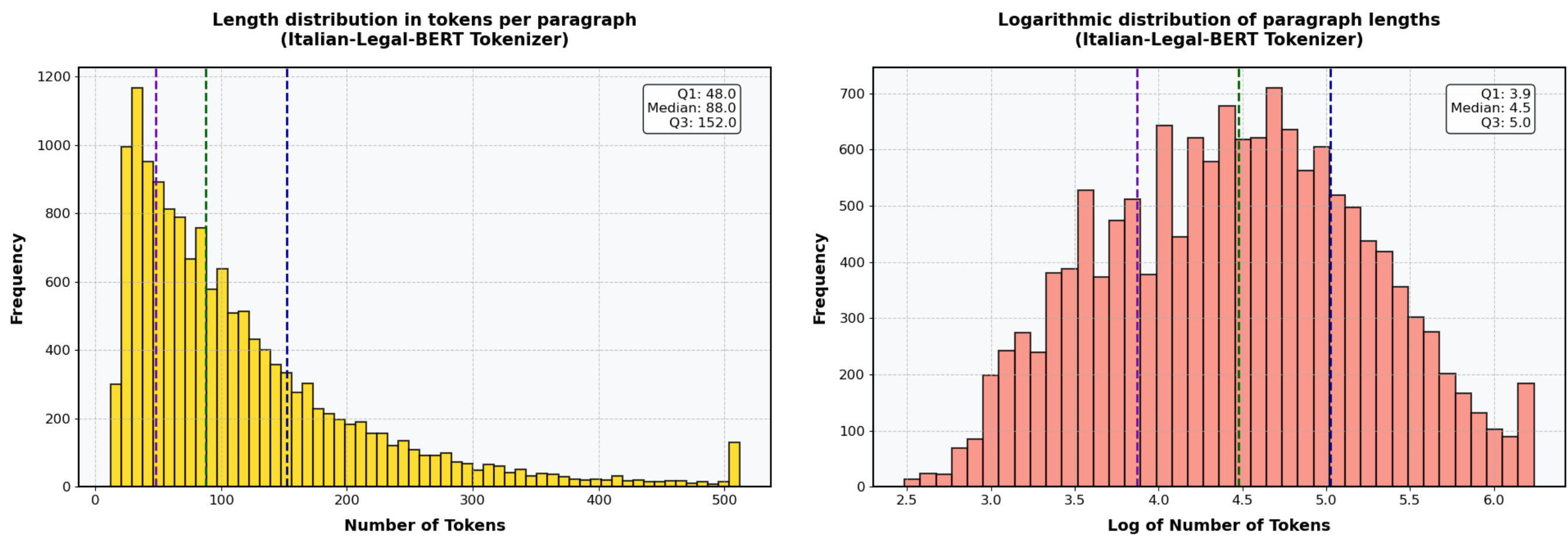}
    \caption{Paragraph length distributions in the analyzed documents. Left: absolute frequency distribution of paragraph lengths tokenized with distil-italian-legal-bert. Right: logarithmic distribution of lengths, showing a more normalized pattern. The dotted lines indicate the first quartile (purple), median (green) and third quartile (blue) of the distribution, respectively.}
    \label{fig:text_lenghts_dist}
\end{figure}

The filtered dataset retains high-quality paragraphs, specifically those in the right-hand tail of the logarithmic distribution of tokenized paragraph lengths, ensuring a focus on content-rich text, Figure \ref{fig:text_lenghts_dist}. Both datasets underwent anonymization to ensure compliance with privacy constraints.

For each version, we conducted a series of experiments to assess topic modeling performance. In particular, we varied the number of topics K between 2 and 50, and computed two key metrics: topic diversity and topic coherence, as defined by the BERTopic framework.

During the experiments, we used the distil-ita-legal-bert embedding model as the backbone for BERTopic. When applied to the dataset generated through our segmentation pipeline, BERTopic demonstrated stable performance across a range of topics from K = 2 to 15. Both topic diversity and coherence remained within desirable ranges, often approaching the upper bounds of the respective metrics and exhibiting a strong positive correlation.

In contrast, when BERTopic was applied to the unsegmented dataset—produced solely through the OCR module—the model frequently generated topics with excessively high diversity values, often surpassing the optimal threshold. Although topic coherence entered the acceptable range starting from K = 18, it tended to decline as the number of topics increased, suggesting reduced interpretability and topic quality at higher values of K.

\begin{figure}[!h]
    \centering
    \includegraphics[width=1\linewidth]{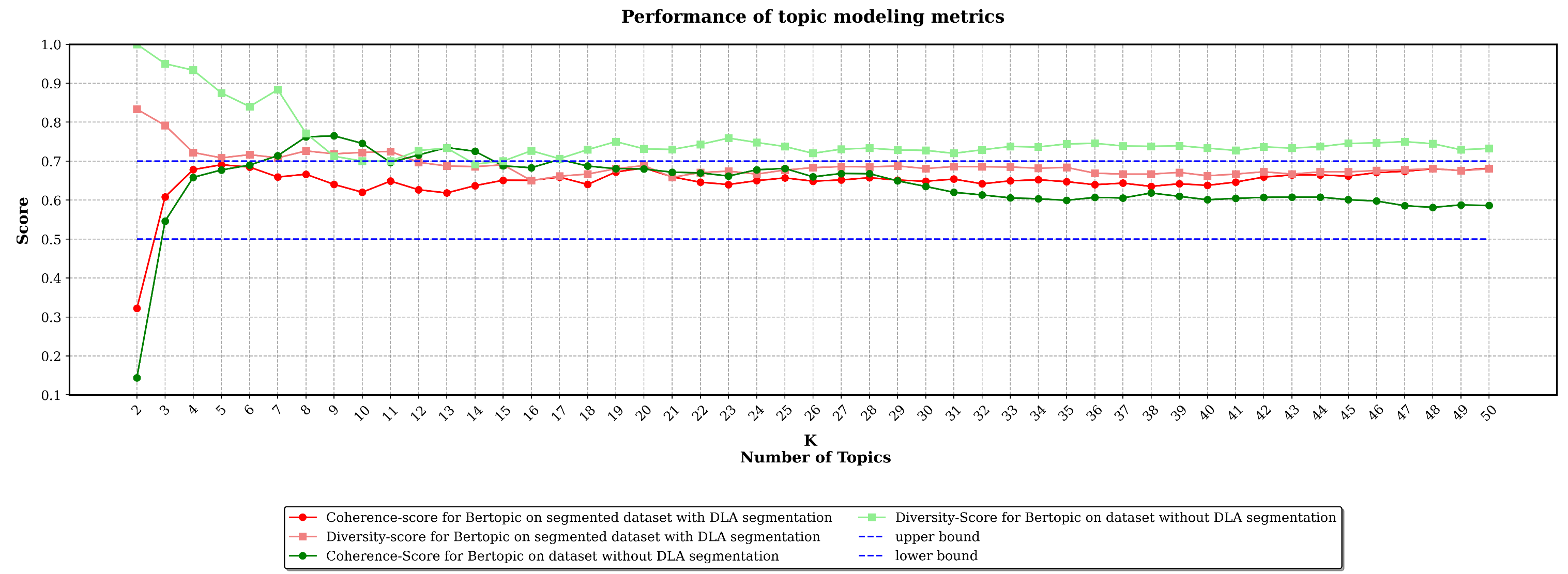}
    \caption{The Figures illustrates the performance trend of BERTopic as a function of the number of topics, using the distil-ita-legal-bert embedding model. The dashed blue lines represent the search boundaries for topic diversity and topic coherence values. The dark red (topic coherence) and light red (topic-diversity) lines represent BERTopic performance on the dataset created with our pipeline, while the dark green (topic coherence) and light green (topic-diversity) lines represent BERTopic performance on data obtained without the use of our pipeline.}

    \label{fig:topic_modeling_cv}
\end{figure}

These findings indicate that our segmentation strategy plays a critical role in improving the quality of topic modeling. By structuring the dataset around coherent and content-rich textual units, the pipeline enables BERTopic to discover a greater number of diverse and consistent topics, enhancing both the granularity and reliability of the extracted topic structure, Figure \ref{fig:topic_modeling_cv}.

To further validate our approach, we repeated the experiment using another general-purpose Italian embedding model, sentence-bert-base-italian-uncased \cite{sentence-bert-base-italian-uncased}. Once again, the dataset generated through our segmentation pipeline yielded superior results, confirming the robustness and generalizability of the method across different embedding strategies.

A detailed summary of the experimental outcomes is presented in Table \ref{tab:topic_metrics}, highlighting the consistent performance gains achieved through structured preprocessing.

\begin{table}[h]
\centering
\setlength{\tabcolsep}{5pt}
\renewcommand{\arraystretch}{1.2}
\begin{tabular}{l|c|c|c|c}
\hline
\multirow{1}{*}{\textbf{Embedding model}} & \textbf{DLA Seg.} & \textbf{K} $\uparrow$ & \textbf{TD} $\uparrow$ & \textbf{$C_{v}$} $\uparrow$ \\
\hline
distil-ita-legal-bert & Yes & 48 & 0.6803 & 0.6809 \\ 
distil-ita-legal-bert & No & 15 & 0.7 & 0.6878 \\ 
sentence-bert-base-italian-uncased & Yes & 47 & 0.6956 & 0.6506 \\ 
sentence-bert-base-italian-uncased & No & 12 & 0.6061 & 0.6280 \\ 
\hline
\end{tabular}
\caption{Comparison of topic modeling results using different versions of BERTopic with and without DLA segmentation.}
\label{tab:topic_metrics}
\end{table}

\subsubsection{Text anonymization evalutations}

Unlike the DLA and OCR modules, the GLiNER-based anonymization component currently lacks a labeled dataset for quantitative performance evaluation. Building such a dataset—following the methodology used for the other components of the pipeline—would require a substantial investment of time and resources, involving the manual annotation of a large corpus of legal documents and the precise definition of sensitive entity types.

To address this challenge, we adopted a pragmatic approach, leveraging a pre-trained GLiNER template available on Hugging Face, specifically oriented toward the Italian language and the detection of Personally Identifiable Information (PII).

This strategy enabled us to rapidly deploy a functional anonymization solution, ensuring practical utility within our pipeline. However, we acknowledge the limitations inherent in this approach, particularly the lack of quantitative validation, which constrains the ability to measure performance systematically.

Despite the absence of a quantitative evaluation framework, we observed that data anonymization has a substantial impact on topic modeling outcomes. In particular, applying BERTopic to non-anonymized data often results in the identification of keywords and representative documents that differ markedly from those produced using anonymized input.

\begin{figure}
    \centering
    \includegraphics[width=0.65\linewidth]{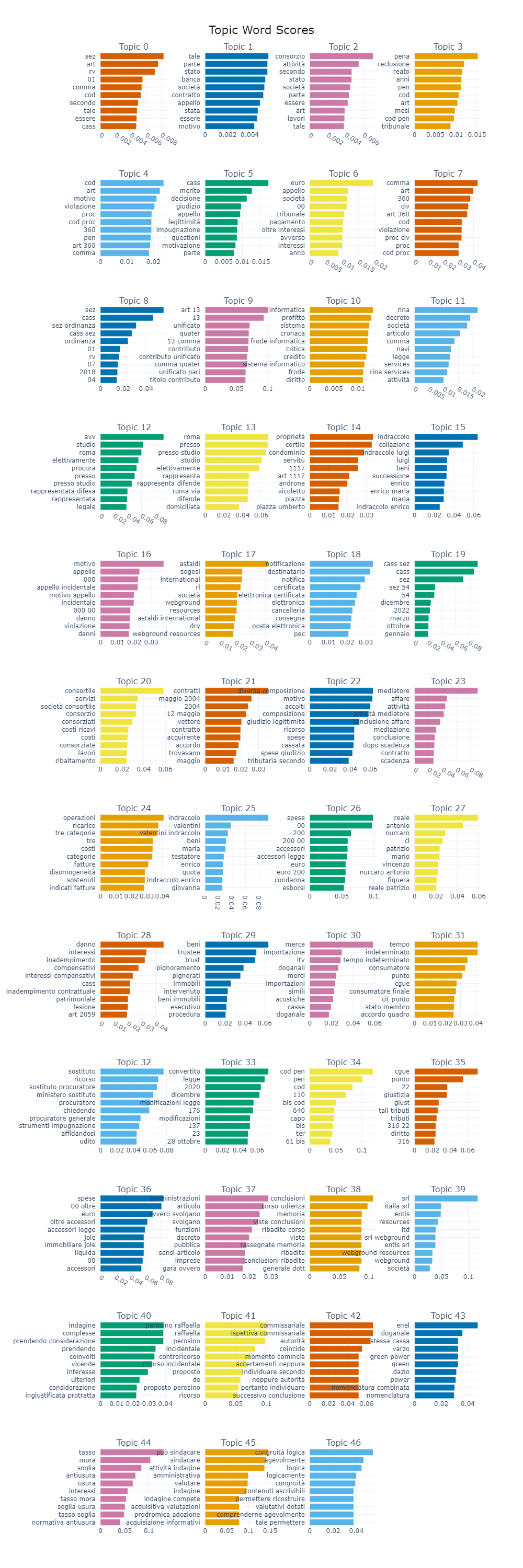}
    \caption{This graph shows the distribution of words describing the topics for each topic. The distributions are displayed as barplots. This case is for the dataset with the non-anonymised text}
    \label{fig:barplot_topic_text_no_anonymized}
\end{figure}

Without anonymization, the model tends to elevate named entities—such as people, locations, and companies to the status of informative keywords, since they frequently appear in the corpus and are interpreted as semantically meaningful, see Figure \ref{fig:barplot_topic_text_no_anonymized}. As a result, many topics are defined around these sensitive identifiers, rather than around the actual substance of the legal text.

\begin{figure}
    \centering
    \includegraphics[width=0.65\linewidth]{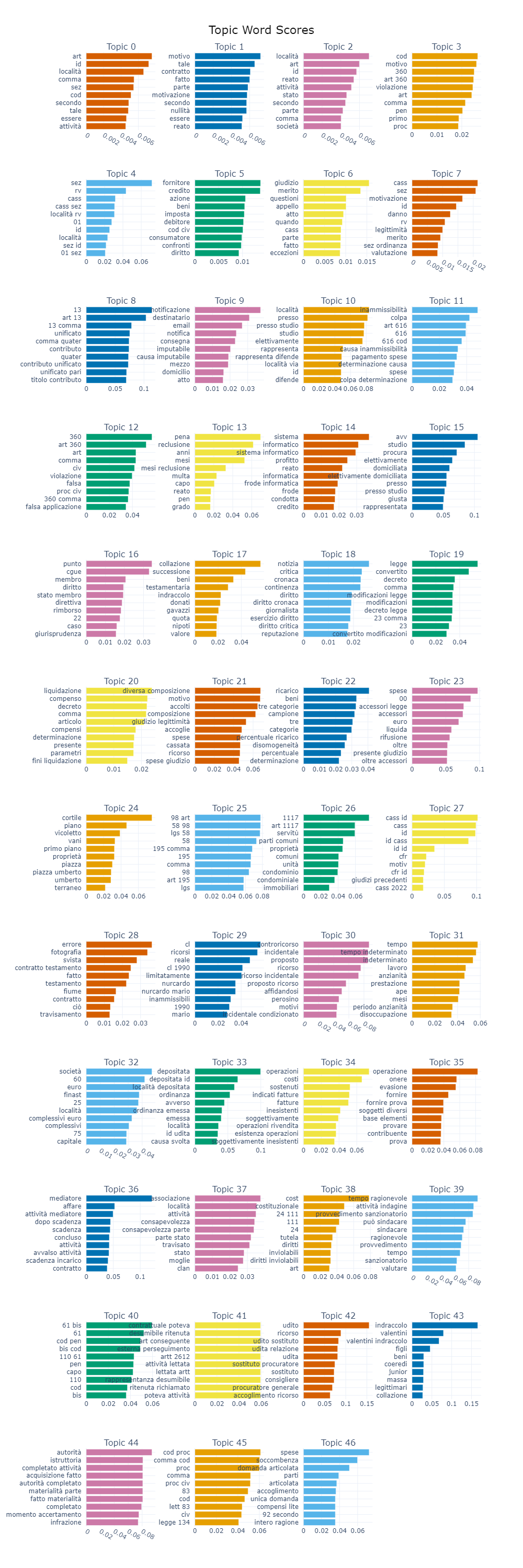}
    \caption{This graph shows the distribution of words describing the topics for each topic. The distributions are displayed as barplots. This case is for the dataset with the anonymised text}
    \label{fig:barplot_topic_text_anonymized}
\end{figure}

By contrast, when the data is anonymized through GLiNER, these entities are replaced with generic placeholders (e.g., $\langle \text{PERSONA}\rangle$, $\langle \text{ORGANIZAZZIONE}\rangle$, $\langle \text{LOCALIT\`A}\rangle$). Because these placeholders recur across multiple documents, their c-TF-IDF scores are low, and BERTopic does not consider them topic-defining. Instead, the model focuses on semantically richer and contextually relevant terms, as illustrated in Figure \ref{fig:barplot_topic_text_anonymized}.

This observation confirms the dual benefit of anonymization: it safeguards sensitive information while simultaneously improving the semantic clarity and interpretability of the topics discovered.

\subsection{Analysis of the detected topics from a legal perspective}
The analysis of the Supreme Court's rulings revealed a number of recurring legal topics pertaining to both procedural law (civil and criminal) and different areas of substantive law.
The main topics identified are examined below, retaining the original numerical identifier for each and summarizing its legal content in a manner consistent with the doctrinal and jurisprudential tradition.
 Normative references are given where appropriate, so as to contextualize each topic with reference to the current legal framework.
\paragraph{Topic 0 - Appeal to the Court of Cassation on Evidence and Reasoning}
Concerns appeals to the Court of Cassation based on errors in the evaluation of evidence or in the reasoning of judgments on the merits. In particular, it refers to the case where the appellate court has failed to examine a decisive fact in the dispute (ground of appeal provided for in Article 360 No. 5 of the Code of Civil Procedure). The Supreme Court can annul the appealed judgment if it finds serious flaws in the reasoning (missing or illogical reasoning) or violations of procedural rules, but it cannot freely review the merits of the case: that is, it does not go into the merits of the evidence assessed by the judges of first and second instance, unless there are manifest errors. In summary, the issue highlights the limits of the Supreme Court's legitimacy review of judgments on the merits, distinguishing errors of law (which can be censured in the Supreme Court) from evaluations of fact (which remain unquestionable).

\paragraph{Topic 1 - Surety bonds and void contract clauses} 
Addresses the validity and interpretation of clauses in surety (guarantee for others' debts) contracts, with attention to potentially void clauses and the limits of review in Supreme Court on contractual interpretation. It is clarified that if some clauses in a contract are found to be void (e.g., standard anti-competitive clauses in bank surety contracts), the court must consider whether the rest of the contract can remain valid without them (partial nullity, Art. 1419 Civil Code). The principle is reiterated that the interpretation of the contract is a factual determination reserved for the court of merit (court or court of appeals); the Supreme Court can intervene only in the case of a clear violation of the legal criteria of interpretation (Art. 1362 et seq. Civil Code) or non-existent/illogical reasoning. In other words, the Supreme Court does not uphold the appeal just because the appellant proposes a different interpretation of the contract, but intervenes if the judgment on the merits ignored the rules of contractual hermeneutics or if its explanation is so deficient as to constitute an error of law.

\paragraph{Topic 6 - Appeals in Tax Litigation: new grounds and “self-sufficiency”}
Concerns procedural rules specific to tax litigation and, more generally, to the Supreme Court. The first principle is the prohibition of new exceptions on appeal: in tax litigation, Article 57 of Legislative Decree 546/1992 prohibits the introduction in the second instance of issues or exceptions that were not raised in the first instance. This means that in tax appeals the parties may discuss only the points already dealt with in the first instance, without adding new grounds for dispute. The second principle is that of self-sufficiency of the appeal to the Supreme Court: a person appealing to the Supreme Court must file an appeal that contains within it all the elements necessary to understand the issue, including the relevant parts of the previous court documents. In particular, if the appellant complains that the trial judges did not consider a certain exception or evidence, the appeal must indicate where and how that issue was raised in the previous stages (e.g., cite the notice of appeal or the record of the hearing in which it was raised). If the appellant raises a new issue in the Supreme Court that does not appear in the appealed judgment or in the previous acts, the Supreme Court will declare it inadmissible. In summary, in the tax process (but the principle applies in general) one cannot change the argument or add new issues at the last moment, and in Cassation the appeal must be valid by reporting everything necessary to assess the errors of law complained.

\paragraph{Topic 7 - Reasoning of Judgments and Limits of Review in the Supreme Court (Press Defamation)}
This topic highlights the extent to which the Supreme Court can review the reasoning of a judgment on the merits, with reference to cases of press defamation. According to Article 111 of the Constitution, the judgment must be reasoned: however, the Supreme Court only verifies compliance with a “constitutional minimum” of reasoning.
This means that the Supreme Court will annul the appealed decision only if the appellate judgment lacks motivation, or if the motivation is merely apparent, irremediably contradictory, or so illogical as to be incomprehensible. In other cases, the assessment of facts and evidence remains reserved for the trial court. In the context of defamation by the press (e.g., an offensive newspaper article), this principle implies that aspects such as reconstructing the facts, evaluating the phrases found to be defamatory, and ascertaining whether the author of the article could invoke the exemption of the right to report or criticism (i.e., writing out of a duty to report without defaming) are considered factual determinations left to the judges of the merits. 
The Supreme Court cannot question these assessments if they are supported by logical reasoning. Its review is limited to whether the judges took into account the basic criteria (e.g., were the incriminated sentences relevant to the public interest of the news story? Were the facts narrated true? Was the language measured?) and that they explained their conclusions consistently. In short, the Supreme Court does not decide whether an article is defamatory or not on the merits, but it checks that the judgment on the merits is reasonably reasoned and that essential elements of assessment are not missing altogether.

\paragraph{Topic 14 - Computer Fraud and Computer System Misuse and Access}
Covers the criminal law of information technology, distinguishing various cases of computer crime. It starts with the difference between computer fraud (Article 640-ter of the Criminal Code) and misuse of stolen payment cards. It is explained that when someone uses a stolen credit or ATM card to withdraw money, it might look like fraud, but technically it falls under a different crime (regulated by Art. 55 co.9 Legislative Decree 231/2007, concerning the use of other people's payment cards) if there is no actual “hacking” of the system. Computer fraud, on the other hand, occurs when the deception is aimed directly at a computer system: for example, manipulating software, introducing false data or altering an electronic system to obtain an undue profit, rather than deceiving a specific person.
The topic emphasizes that in computer fraud the deception affects the functionality of the system (the computer, the server, the ATM), while traditional fraud or misuse of cards affects the human victim who is misled. In addition, the crime of abusive access to a computer system (Art. 615-ter of the Criminal Code) is examined, pointing out that it can be committed not only by those who are completely outside the system, but also by those who were initially authorized to access it: for example, an employee of a company who has credentials for the company system but uses them beyond the permitted limits or for purposes other than those authorized. In practice, if a person enters a protected computer system by violating the conditions set by the owner (perhaps by accessing confidential data without permission while having an account), he or she may still be liable for abusive access. In summary, the topic sheds light on the different types of computer crime, explaining the subtle differences: using a stolen card vs. computer fraud vs. abusive access, and how case law defines the boundaries between these cases.
\paragraph{Topic 26 - Easements over condominium property and “condominium” status}
Concerns a question of real estate law: whether the holder of an easement over a common part of a condominium should be considered a condominium owner with rights and duties equal to the other owners of the building. A servient easement is, for example, the right of way over someone else's driveway, or the right to use a yard to run pipes, in favor of a neighboring property. In this case, let us imagine that a person from outside the condominium has the right to pass through the hallway or courtyard of a condominium to access his or her property: this person does not own anything in the condominium, he or she only has a right of easement. The legal question is: does having this easement make her a condominium owner (i.e., part of the condominium)? The answer, clarified by the Supreme Court, is negative. The holder of an easement over condominium property does not automatically become a condominium owner and therefore does not acquire co-ownership of that property nor is he or she required, as a rule, to contribute to general condominium expenses. The condominium owner's rights (e.g., voting in the assembly, or owning a share of the common parts such as stairs, roof, etc.) derive from ownership of a property unit in the building, not from a mere right of easement.
Similarly, a condominium owner's obligations (such as paying expenses for elevator maintenance, roof maintenance, etc.) do not apply to those outside the condominium property. In our example, the holder of an easement of passage will only contribute to the specific expenses associated with the exercise of that easement, if any (e.g., he or she may have to contribute to keeping the driveway he or she uses in good condition, according to Article 1069 of the Civil Code), but he or she will not pay the expenses for the elevator or the stair light because he or she is not a condominium owner. In summary, the recognition of an easement in favor of an outsider does not bring him into the “condominium”: he remains a third party who enjoys a limited right, governed by the rules on easements, and is not part of the condominium community unless otherwise agreed by contract.

\paragraph{Topic 28 - Appeal to the Supreme Court for trial error}
In Italian civil law, “misrepresentation of evidence” refers to when a judge misunderstands evidence, basing the decision on something other than what that evidence actually indicates. A distinction is made between errors of perception and errors of interpretation: an error of perception occurs when the judge “missees” the content of evidence (e.g., misreading a date on a document, or mistaking a picture of a car for a picture of a river), while an error of interpretation occurs when the judge correctly understands the evidence but gives it the wrong meaning or weight. This distinction is important because it determines whether and how the error can be rectified. A misrepresentation due to an obvious oversight (factual error) can be corrected by seeking a reversal of the judgment, that is, by reopening the trial to correct that factual error. In contrast, an error in the interpretation of evidence normally cannot be challenged in the Supreme Court, since the Supreme Court does not review the facts. Only in exceptional cases where the error is so serious as to amount to a failure to consider decisive evidence does the law allow an appeal to the Court of Cassation (Article 360, paragraph 1, no. 5 of the Code of Civil Procedure, relating to failure to examine a decisive fact) to correct the misrepresentation.

\begin{figure}[!h]
    \centering
    \includegraphics[width=0.9\linewidth]{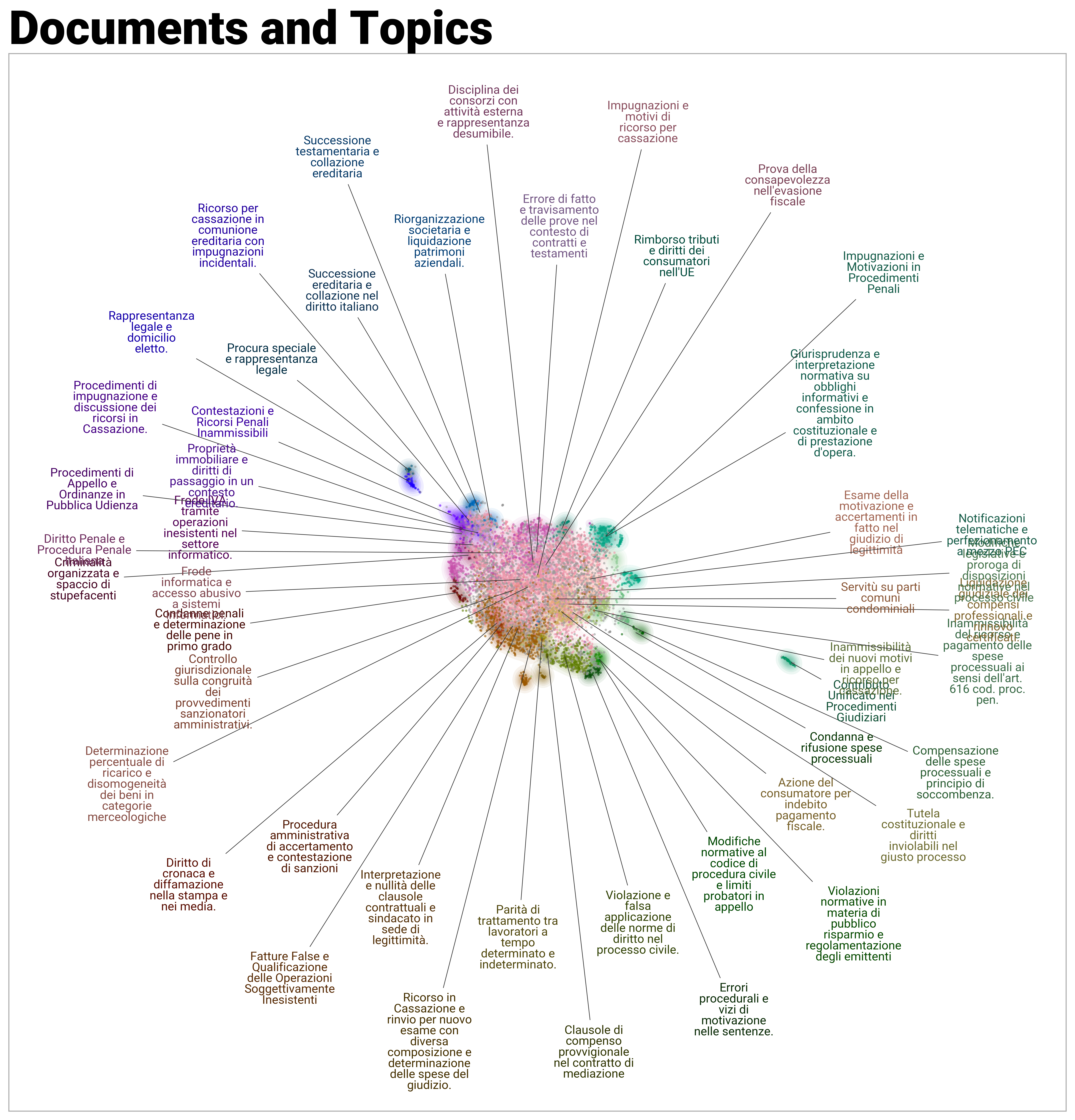}
    \caption{Documents and Topics in Italian Legal Procedures: A 2D scatter plot showing the complex dataset structure. The color-coded clusters represent different legal topic (administrative law, criminal procedure, digital authentication, and COVID-19 related regulations).}
    \label{fig:scatter_plot_topic}
\end{figure}

We conclude this subsection by presenting Figure \ref{fig:scatter_plot_topic}, which visualizes all detected topics within the 2D embedding space of the documents in the dataset.

\subsubsection{LLMs evalutations}

After analyzing the topic structure of our legal corpus, we investigated the ability of LLMs to interpret and describe the topics identified by BERTopic. Our focus was on assessing their performance in two key tasks: generating descriptive labels and producing concise summaries, which we compared against content produced by a legal domain expert.

To establish a reference standard, we collaborated with a legal expert who manually interpreted and summarized each topic. This allowed us to carry out a dual evaluation—both quantitative and qualitative—of the outputs produced by the LLMs.

In the label generation task, all models achieved comparable results, with LLaMA 3.1 (8B) performing slightly better, attaining an F1 score of 0.8256. However, the differences between models were minor, suggesting similar effectiveness in producing accurate topic labels.

 Although the LLAMA model achieved the highest score for label generation in terms of F1, we saw its outputs and were not satisfied because the model frequently ignored the prompt by providing a summary rather than a short label. Claude and GPT's outputs, on the other hand, in addition to being semantically correct, turned out to be true labels.

For the summary generation task, Claude 3.7  was the top performer, with an F1 score of 0.9130, marginally surpassing GPT-4o. Both models generated high-quality summaries, clearly outperforming LLaMA 3.1, which often produced generic or less insightful content. These results are expected, given the training scale and capabilities of Claude and GPT-4o.

\begin{figure}[!h]
    \centering
    \includegraphics[width=0.75\linewidth]{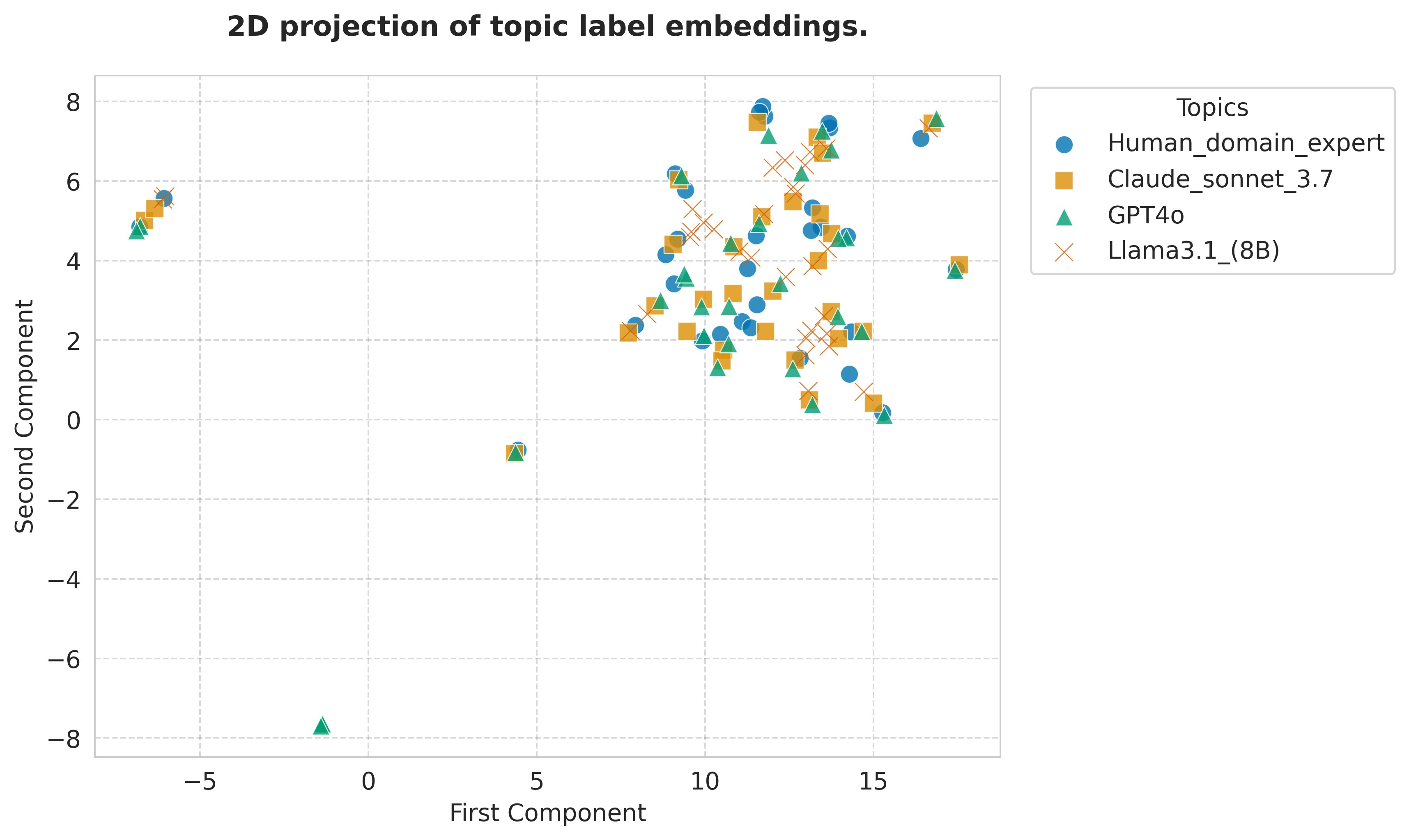}
    \caption{Two-dimensional projection of topic label embeddings generated by different language models (Claude-Sonnet 3.7, GPT4o, Llama 3.1 (8B)) and a human expert. The visualization shows significant spatial overlap between the labels generated by the different LLMs, suggesting strong semantic agreement.}
    \label{fig:2D_projection_of_topic_label_embeddings}
\end{figure}

\begin{figure}[!h]
    \centering
    \includegraphics[width=0.75\linewidth]{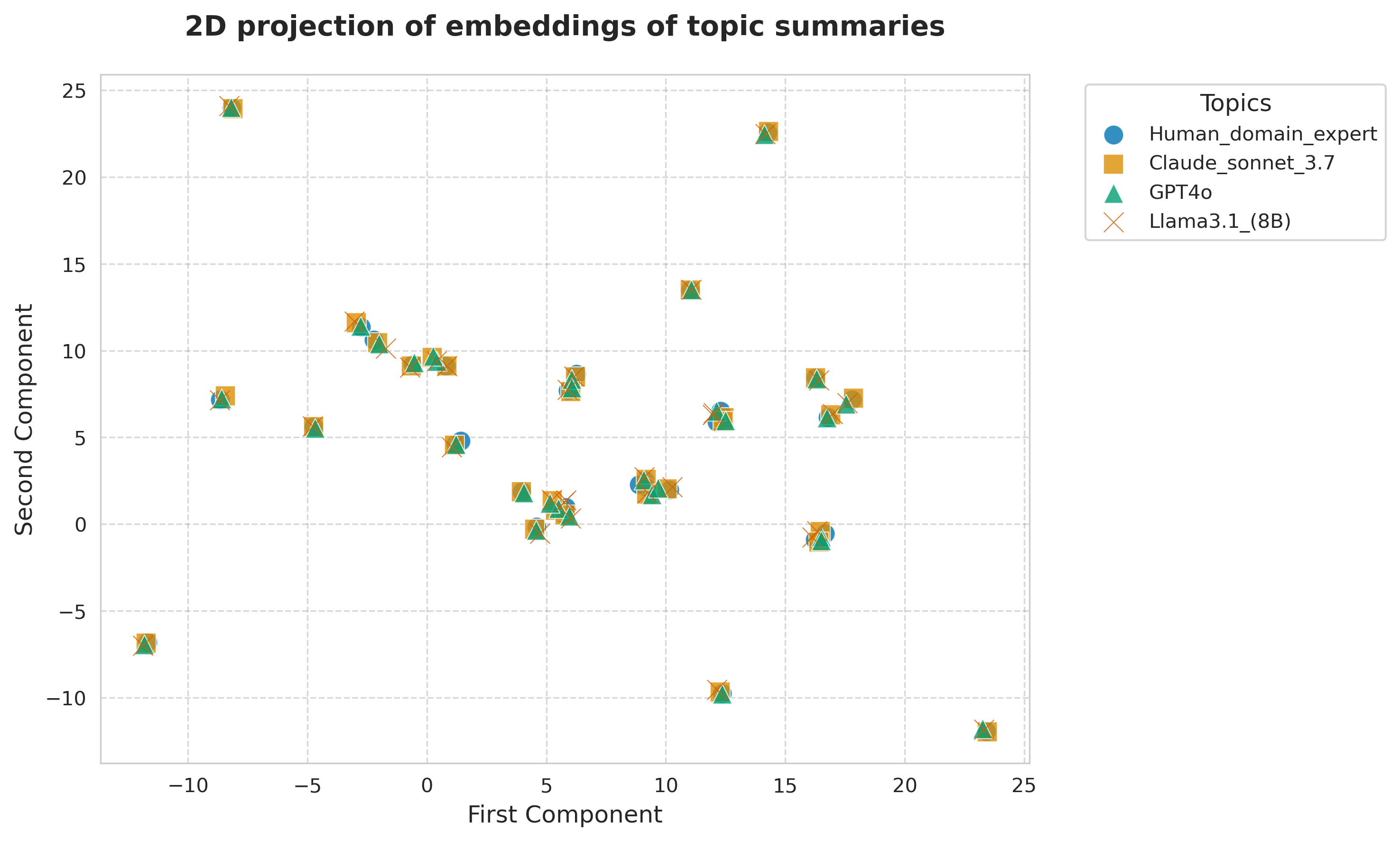}
    \caption{Two-dimensional projection of topic text summary embeddings generated by different language models (Claude-Sonnet 3.7, GPT4, Llama 3.1(8B)) and a human expert. The visualization shows significant spatial overlap between the summury generated by the different LLMs, suggesting strong semantic agreement.}
    \label{fig:2D_projection_of_topic_summary_embeddings}
\end{figure}

In addition to this quantitative evaluation (see Table \ref{tab:merged_results}), we conducted a qualitative analysis by embedding all labels and summaries—including those generated by the expert—and projecting them into 2D space using UMAP. The resulting visualizations (Figures \ref{fig:2D_projection_of_topic_label_embeddings} and \ref{fig:2D_projection_of_topic_summary_embeddings}) show that the LLM outputs cluster closely around expert-generated content, indicating strong semantic alignment.

\begin{table}[h]
\centering
\setlength{\tabcolsep}{5pt}
\renewcommand{\arraystretch}{1.2}
\begin{tabular}{l|ccc|ccc}
\hline
\multirow{1}{*}{\textbf{Model}} & \multicolumn{3}{c|}{\textbf{Topic-Label Generation}} & \multicolumn{3}{c}{\textbf{Topic Summarization}} \\
 & \textbf{Prec.} $\uparrow$ & \textbf{Rec.} $\uparrow$ & \textbf{F1} $\uparrow$ & \textbf{Prec.} $\uparrow$ & \textbf{Rec.} $\uparrow$ & \textbf{F1} $\uparrow$ \\
\hline
Claude-Sonnet3.7 & 0.8220 & 0.8046 & 0.8119 & 0.9058 & \textbf{0.9205} & \textbf{0.9130} \\
GPT4 &  0.7895 & \textbf{0.8153} & 0.7995 & 0.7995 & 0.9180 & 0.9122 \\
Llama3.1 (8B) & \textbf{0.8558} & 0.8010 & \textbf{0.8256} & \textbf{0.9092} & 0.8817 & 0.8951 \\
\hline
\end{tabular}
\caption{Comparison of BERTScore metrics for Topic-Label Generation and Topic Summarization tasks across different models. Best results are highlighted in \textbf{bold}.}
\label{tab:merged_results}
\end{table}

Below, Tables [\ref{tab:topic_comparison_4}, \ref{tab:topic_comparison_5}, \ref{tab:topic_comparison_6}, \ref{tab:topic_comparison_7}, \ref{tab:topic_comparison_8}, \ref{tab:topic_comparison_9} and \ref{tab:topic_comparison_10}] present a selection of labels and summaries generated by the LLMs and the domain expert for reference and comparison.

\begin{table}[ht]
\centering
\caption{Comparison of model-generated labels and summaries for topic 0}
\label{tab:topic_comparison_4}
\begin{tabular}{|l|p{4cm}|p{7.5cm}|}
\hline
\textbf{Source} & \textbf{Label} & \textbf{Summary} \\
\hline
\textbf{Expert} & Ricorso in Cassazione per valutazione della prova e motivazione sentenze merito & Il presente topic fa riferimento a questioni procedurali relative all'impugnazione delle sentenze di merito in Cassazione, alla valutazione della prova e alla motivazione delle decisioni dei giudici di merito. In particolare, si tratta dell'ammissibilità del ricorso per Cassazione ai sensi dell'art. 360, n. 5, relativo all'omesso esame di fatti decisivi, alla motivazione della sentenza impugnata e alla corretta applicazione delle norme processuali. \\
\hline
\textbf{Claude} & Ricorso per omesso esame di fatti decisivi in Cassazione & Il topic riguarda i ricorsi in Cassazione ai sensi dell'art. 360 del codice di procedura civile, con particolare riferimento al motivo di cui al comma 1, n. 5 (omesso esame di fatto decisivo). I documenti analizzano i requisiti di ammissibilità di tali ricorsi, le condizioni in cui sono considerati inammissibili, e la giurisprudenza della Corte di Cassazione in materia. Vengono trattate anche questioni relative alla violazione di norme procedurali (come gli artt. 115, 116 e 132 c.p.c.) e all'onere della prova (art. 2697 c.c.), evidenziando i limiti del sindacato di legittimità sulle valutazioni di merito operate dai giudici di secondo grado. Il tema si inserisce nel contesto del controllo di legittimità delle sentenze e della distinzione tra errori di diritto censurabile e valutazioni di merito non sindacabili in sede di legittimità. \\
\hline
\textbf{GPT-4} & Impugnazioni e motivi di ricorso per cassazione & Il tema analizzato riguarda le modalità e i limiti del ricorso per cassazione nel diritto processuale civile italiano, in particolare ai sensi dell'articolo 360 del codice di procedura civile. Sono trattati gli aspetti relativi all'ammissibilità dei ricorsi per omesso esame di fatti decisivi e violazioni procedurali, nonché la corretta interpretazione delle disposizioni normative applicabili in sede di giudizio di appello e di cassazione. Vengono esaminate diverse causali di inammissibilità del ricorso, come la mancata indicazione delle ragioni di fatto poste a base delle decisioni impugnate, e si discutono gli oneri probatori e i criteri di valutazione delle prove, tra cui il prudente apprezzamento del giudice. Il topic esplora anche le interpretazioni giurisprudenziali su questi temi e l'influenza delle sentenze di sezioni unite della Corte di Cassazione, sottolineando la complessità dei giudizi in secondo grado e dei procedimenti di impugnazione. \\
\hline
\textbf{LLaMA} & Analisi di un ricorso per cassazione presentato da un'organizzazione contro una sentenza di rigetto dell'appello, con motivi di censura relative alla violazione di norme processuali e civilistiche & Analisi di un ricorso per cassazione presentato da un'organizzazione contro una sentenza di rigetto dell'appello, con motivi di censura relative alla violazione di norme processuali e civilistiche, tra cui omesso esame di fatti decisivi, violazione dell'art. 132 cod. proc. civ. e dell'art. 36, comma secondo, n. 4, del decreto legislativo, e falsa applicazione dell'art. 115 cod. proc. civ. \\
\hline
\end{tabular}
\end{table}

\begin{table}[ht]
\centering
\caption{Comparison of model-generated labels and summaries for topic 1}
\label{tab:topic_comparison_5}
\begin{tabular}{|l|p{4cm}|p{7.5cm}|}
\hline
\textbf{Source} & \textbf{Label} & \textbf{Summary} \\
\hline
\textbf{Expert} & Fideiussione e interpretazione contrattuale & Il presente topic ha ad oggetto  l'interpretazione e la validità delle clausole nei contratti di fideiussione, con particolare riferimento alla nullità parziale o totale del contratto e ai criteri di ermeneutica contrattuale. Inoltre, viene trattato il problema dei limiti del sindacato di legittimità in Cassazione sull'interpretazione dei contratti, ammissibile solo se vi è una violazione manifesta dei criteri legali di ermeneutica contrattuale. \\
\hline
\textbf{Claude} & Interpretazione contrattuale e nullità delle clausole negoziali in sede giudiziale & Interpretazione contrattuale e nullità delle clausole nel diritto italiano, con particolare riferimento al sindacato di legittimità in sede di appello. Il tema si concentra sull'accertamento della volontà delle parti come questione di fatto riservata al giudice di merito, i limiti del ricorso in Cassazione per violazione dei canoni ermeneutici (artt. 1362 e ss. c.c.), i requisiti per censurare l'interpretazione contrattuale in sede di legittimità, e l'applicazione dell'art. 1419 c.c. sulla nullità parziale del contratto in presenza di clausole nulle, con specifico riferimento al caso di clausole dichiarate nulle in contratti di fideiussione.\\
\hline
\textbf{GPT-4} & Interpretazione e nullità delle clausole contrattuali e sindacato in sede di legittimità. & Il topic riguarda questioni giuridiche relative alla nullità parziale o totale dei contratti, con particolare attenzione alle fideiussioni e alle clausole contrattuali dichiarate nulle. Si discute il rapporto tra la nullità di singole clausole e la validità del contratto nel suo complesso, secondo l'art. 1419 del codice civile. Inoltre, viene approfondito il tema dell'interpretazione delle volontà contrattuali delle parti, che è considerato un accertamento di fatto riservato al giudice del merito. Le decisioni giudiziarie sulla validità e interpretazione dei contratti sono sindacabili in Cassazione solo per vizi di motivazione, rispetto ai criteri legali d'interpretazione contrattuale previsti dagli articoli 1362 e seguenti del codice civile. La contrapposizione di un'interpretazione alternativa del ricorrente a quella accolta dalla sentenza impugnata non basta per una censura valida, a meno che ci siano considerazioni illogiche o insufficienti. \\
\hline
\textbf{LLaMA} & La discussione su come debba essere interpretato un contratto di fideiussione in presenza di clausole dichiarate nulle, e le condizioni per la censura di tale interpretazione in sede di & La discussione su come debba essere interpretato un contratto di fideiussione in presenza di clausole dichiarate nulle, e le condizioni per la sindacabilità in sede di legittimità dell'accertamento della volontà delle parti in relazione al contenuto di un negozio giuridico. \\
\hline
\end{tabular}
\end{table}

\begin{table}[ht]
\centering
\caption{Comparison of model-generated labels and summaries for topic 6}
\label{tab:topic_comparison_6}
\begin{tabular}{|l|p{4cm}|p{7.5cm}|}
\hline
\textbf{Source} & \textbf{Label} & \textbf{Summary} \\
\hline
\textbf{Expert} & Ricorso in Cassazione nel processo tributario per i limiti alle eccezioni e al principio di autosufficienza & Il presente topic ha ad oggetto la materia di contenzioso tributario e ricorso per cassazione, in particolare il divieto di proporre nuove eccezioni in appello e sull'autosufficienza del ricorso. Sulla base del D.Lgs. n. 546/1992, come interpretato dalla Cassazione, in appello non si possono proporre nuove eccezioni in senso tecnico. In Cassazione, il ricorso deve basarsi su questioni già affrontate in appello, non essendo ammessa l'introduzione di nuovi oggetti di indagine. Per il principio di autosufficienza, se con il ricorso in Cassazione sono prospettate questioni prive di cenno nella sentenza impugnata, il ricorso deve allegare la loro deduzione dinanzi al giudice di merito e indicare in quale specifico atto del giudizio precedente lo abbia fatto. \\
\hline
\textbf{Claude} & La disciplina delle eccezioni e dei nuovi argomenti nei gradi di giudizio nel diritto processuale italiano & Il topic tratta delle regole procedurali che disciplinano la presentazione di eccezioni, domande e questioni nei diversi gradi di giudizio nel sistema giuridico italiano, con particolare attenzione ai limiti alla proposizione di nuove eccezioni in appello e in Cassazione. Emerge in particolare il principio di inammissibilità delle questioni nuove nei gradi successivi al primo: in appello vige il divieto di proporre nuove eccezioni (come stabilito dall'art. 57, comma 2, del d.lgs. n. 546), mentre in sede di legittimità (Cassazione) non possono essere prospettate questioni che non siano già state comprese nel tema del decidere del giudizio di appello. Viene inoltre evidenziato il principio di autosufficienza del ricorso per Cassazione, in base al quale, quando vengono sollevate questioni non menzionate nella sentenza impugnata, il ricorrente deve, a pena di inammissibilità, non solo allegare di averle già dedotte nei precedenti gradi di giudizio, ma anche indicare precisamente in quale specifico atto processuale lo abbia fatto.\\
\hline
\textbf{GPT-4} & Inammissibilità dei nuovi motivi in appello e ricorso per cassazione. & Il topic riguarda il processo di appello in ambito giuridico italiano, con particolare riferimento ai vizi delle domande e delle decisioni nei procedimenti legali. Si focalizza su aspetti critici come l'inammissibilità di nuove eccezioni e questioni non dedotte nel giudizio di merito quando si ricorre in cassazione. La normativa pertinente include, tra altri, l'articolo 57 del decreto legislativo n. 546, che disciplina il contenzioso tributario, vietando la proposizione di nuove eccezioni in appello. Inoltre, viene sottolineata l'importanza del principio di autosufficienza del ricorso, che impone di indicare espressamente gli atti processuali precedenti in cui una questione è stata proposta al giudice di merito. Le decisioni devono basarsi su fatti e questioni già introdotti nel grado di appello, evitando di sollevare nel giudizio di legittimità elementi nuovi non esaminati in precedenza. \\
\hline
\textbf{LLaMA} & Le regole per la proposizione di eccezioni e domande in sede di appello e cassazione nel processo civile italiano, con particolare attenzione all'autosufficienza del ricorso e alla inam & Le regole per la proposizione di eccezioni e domande in sede di appello e cassazione nel processo civile italiano, con particolare attenzione all'autosufficienza del ricorso e alla inammissibilità dei motivi nuovi. \\
\hline
\end{tabular}
\end{table}

\begin{table}[ht]
\centering
\caption{Comparison of model-generated labels and summaries for topic 7}
\label{tab:topic_comparison_7}
\begin{tabular}{|l|p{4cm}|p{7.5cm}|}
\hline
\textbf{Source} & \textbf{Label} & \textbf{Summary} \\
\hline
\textbf{Expert} & Motivazione della sentenza e limiti del sindacato di legittimità in tema di diffamazione a mezzo stampa & Il topic ha ad oggetto il tema della motivazione della sentenza e del suo sindacato da parte della Cassazione. L'art. 111, comma 6, Cost. viene violato solo qualora la motivazione sia totalmente mancante o meramente apparente, ovvero si fondi su un contrasto irriducibile tra affermazioni inconciliabili, o risulti perplessa ed obiettivamente incomprensibile. Con particolare riferimento alla materia del risarcimento dei danni da diffamazione a mezzo della stampa, la ricostruzione dei fatti, la valutazione del contenuto degli scritti, l'apprezzamento delle espressioni usate come lesive dell'altrui reputazione, la valutazione dell'esistenza o meno dell'esimente dell'esercizio dei diritti di cronaca e di critica costituiscono oggetto di accertamenti in fatto, riservati al giudice di merito ed insindacabili in sede di legittimità. Il controllo affidato alla Cassazione è limitato alla verifica dell'avvenuto esame, da parte del giudice del merito, della sussistenza dei requisiti della continenza, della veridicità dei fatti narrati e dell'interesse pubblico alla diffusione delle notizie, nonché al sindacato della congruità e logicità della motivazione, secondo la previsione dell'art. 360, comma 1, n. 5, c.p.c., restando estraneo al giudizio di legittimità l'accertamento relativo alla capacità diffamatoria delle espressioni in contestazione. \\
\hline
\textbf{Claude} & Limiti al sindacato di legittimità della Cassazione in materia di motivazione e accertamenti di fatto & Limiti al sindacato di legittimità della Cassazione sulla motivazione delle sentenze. Il topic riguarda i limiti del sindacato di legittimità della Corte di Cassazione sulla motivazione delle sentenze, con particolare riferimento alle valutazioni di merito operate dai giudici di grado inferiore. Emerge una giurisprudenza consolidata secondo cui: 1. Il controllo della Cassazione sulla motivazione è circoscritto alla verifica del "minimo costituzionale" richiesto dall'art. 111, comma 6, Cost., che risulta violato solo in caso di motivazione totalmente mancante, meramente apparente, fondata su contrasti irriducibili o oggettivamente incomprensibile.2. In specifici ambiti, come nelle azioni di risarcimento danni da diffamazione a mezzo stampa, accertamenti fattuali quali la ricostruzione storica dei fatti, la valutazione del contenuto degli scritti, l'apprezzamento delle espressioni diffamatorie e la valutazione delle esimenti sono riservati al giudice di merito e insindacabili in sede di legittimità se sorretti da argomentata motivazione.3. Il controllo della Corte di Cassazione è limitato alla verifica dell'avvenuto esame, da parte del giudice di merito, della sussistenza dei requisiti necessari (come continenza, verità dei fatti, interesse pubblico) e al sindacato della congruità e logicità della motivazione secondo l'art. 360, comma 1, n. 5 c.p.c.\\
\hline
\textbf{GPT-4} &Esame della motivazione e accertamenti in fatto nel giudizio di legittimità & Il tema trattato è relativo ai vizi di motivazione e agli errori in decisioni giudiziarie, in particolare nell\'ambito della giurisprudenza italiana. Si esamina il sindacato di legittimità sulla motivazione delle sentenze, sottolineando che esso è limitato a verificare il rispetto del "minimo costituzionale" come richiesto dall\'articolo 111 della Costituzione. Questo controllo si applica solo quando la motivazione è totalmente assente, meramente apparente, o presenta contraddizioni irriducibili. Inoltre, il tema coinvolge la valutazione di casi di risarcimento del danno da diffamazione a mezzo stampa, con particolare riferimento all\'apprezzamento dei fatti e delle espressioni usate. La giurisprudenza sottolinea che tali accertamenti sono riservati al giudice di merito e non sono sindacabili in sede di legittimità se basati su motivazioni argomentate. Il ruolo della legittimità è quindi limitato a verificare la congruità e la logicità della motivazione, secondo l\'articolo 360, comma 1, numero 5, del codice di procedura civile, restando estraneo l\'accertamento della capacità diffamatoria delle espressioni contestate. \\
\hline
\textbf{LLaMA} & La discussione verte sulle regole di accertamento e valutazione dei fatti in materia di danno e diffamazione, con particolare attenzione ai limiti del sindacato di legittimità sulla & La discussione verte sulle regole di accertamento e valutazione dei fatti in materia di danno e diffamazione, con particolare attenzione ai limiti del controllo di legittimità e all'importanza della motivazione nella decisione del giudice di merito. \\
\hline
\end{tabular}
\end{table}

\begin{table}[ht]
\centering
\caption{Comparison of model-generated labels and summaries for topic 14}
\label{tab:topic_comparison_8}
\begin{tabular}{|l|p{4cm}|p{7.5cm}|}
\hline
\textbf{Source} & \textbf{Label} & \textbf{Summary} \\
\hline
\textbf{Expert} & Frode informatica e accesso abusivo a sistema informatico & Il presente topic riporta il tema della distinzione tra il reato di utilizzo indebito di carte di credito o bancomat sottratte a terzi per prelevare denaro (introdocco con l'art. 55, comma 9, del D.Lgs. n. 231/2007)  e il reato di frode informatica (art. 640-ter c.p.), precisando che che, nel caso della frode informatica, il raggiro è rivolto più al sistema informatico che direttamente alla persona fisica o giuridica. Il topic tratta poi dell'accesso abusivo a un sistema informatico (art. 615-ter c.p.), evidenziando come questo reato possa essere  commesso anche da colui che,  pur essendo abilitato, accede o si mantenga in un sistema informatico o telematico protetto, violando le condizioni ed i limiti risultanti dal  complesso delle prescrizioni impartite dal titolare del sistema, mantenendo gli scopi e le finalità che lo avevano motivato a fare ingresso nel sistema. \\
\hline
\textbf{Claude} & La frode informatica e l'uso indebito di carte di credito altrui & Il topic riguarda il reato di frode informatica previsto dall'art. 640-ter del codice penale italiano e le sue caratteristiche distintive rispetto ad altre fattispecie criminose simili. Il focus è sulla qualificazione giuridica della condotta di chi ottiene un ingiusto profitto alterando sistemi informatici o utilizzando indebitamente dati e strumenti di pagamento elettronici come carte di credito. La giurisprudenza analizzata evidenzia i confini tra la frode informatica, che richiede specificamente la manipolazione di un sistema informatico, e altri reati come l'indebito utilizzo di strumenti di pagamento (disciplinato dall'art. 55 del D.Lgs. 231/2007) o l'accesso abusivo a sistemi informatici (art. 615-ter c.p.). Vengono esaminati gli elementi costitutivi del delitto di frode informatica, con particolare attenzione all'alterazione del funzionamento di sistemi informatici quale presupposto necessario per la configurabilità del reato, distinguendolo dai casi in cui l'attività decettiva è rivolta direttamente contro una persona fisica anzichè contro un sistema informatico.\\
\hline
\textbf{GPT-4} & Frode informatica e accesso abusivo a sistemi informatici. & Il tema in questione riguarda il reato di frode informatica, disciplinato dall'articolo 640 ter del codice penale italiano. Questo reato si configura quando un individuo, attraverso la manipolazione di un sistema informatico o telematico, ottiene un ingiusto profitto per sè o per altri. Si tratta di una condotta che non è diretta direttamente contro una persona, ma piuttosto contro l'integrità e la corretta funzionalità di un sistema informatico. Le condotte possono includere l'uso non autorizzato di carte di credito o bancomat rubati, l'ingresso o il mantenimento in un sistema informatico protetto senza autorizzazione o violando le norme di accesso prescritte. Le situazioni descritte nei documenti si riferiscono anche a tecniche di inganno che, attraverso documenti falsi e raggiri, inducono le organizzazioni a compiere atti dispositivi dannosi per la persona offesa. Inoltre, si evidenzia l'importanza della motivazione e della qualificazione legale delle condotte esaminate nei procedimenti giudiziari relativi a tali reati. \\
\hline
\textbf{LLaMA} & Il topic verte sulla descrizione di casi di frode informatica e sui relativi reati commessi tramite sistemi informatici e telematici. I casi in questione coinvolgono l'utilizzo di carte di cred &Il topic verte sulla figura del reato di frode informatica, con particolare riferimento all'utilizzo indebito di sistemi informatici e strumenti di pagamento per ottenere un ingiusto profitto. I documenti presentano casi di condotte illecite commesse da individui o organizzazioni che hanno alterato o manipolato sistemi informatici per trarne vantaggio personale o altrui. La descrizione può essere riassunta come segue: topic: Reati di frode informatica e ingiusto profitto: utilizzo indebito di sistemi informatici e strumenti di pagamento per ottenere un ingiusto profitto \\
\hline
\end{tabular}
\end{table}

\begin{table}[ht]
\centering
\caption{Comparison of model-generated labels and summaries for topic 26}
\label{tab:topic_comparison_9}
\begin{tabular}{|l|p{4cm}|p{7.5cm}|}
\hline
\textbf{Source} & \textbf{Label} & \textbf{Summary} \\
\hline
\textbf{Expert} & Servitù su beni condominiali e qualità di condomino & Il presente topic ha ad oggetto la questione se la titolarità di un diritto di servitù prediale su un bene condominiale faccia acquisire gli stessi diritti ed obblighi di un condomino. Chi ha simile diritto reale su un bene condominiale non è tenuto a contribuire alle spese generali del condominio, a meno che non vi sia un accordo specifico o una previsione di legge che lo imponga. L'esercizio della servitù avviene sulla base degli artt. 1063 e ss. c.c. e art. 1069, comma 3, c.c. Allo stesso modo, il riconoscimento in favore di un'unità immobiliare individuale di un diritto di servitù su una parte comune non vale di per sè nè a conferire nè a negare al relativo titolare la qualità di condomino agli effetti della contitolarità delle parti comuni dell'edificio, dell'attribuzione dei diritti sulle stesse in proporzione al valore della rispettiva unità immobiliare e dell'obbligo di contribuire alle spese per la conservazione e il godimento.  \\
\hline
\textbf{Claude} & Servitù prediali su beni condominiali & Il topic riguarda il regime giuridico delle servitù prediali costituite su beni condominiali (come portoni d'ingresso e androni) a favore di proprietà esclusive esterne al condominio. La questione centrale è che la titolarità di una servitù su parti comuni condominiali non attribuisce automaticamente al proprietario del fondo dominante la qualifica di condomino.La giurisprudenza (in particolare la sentenza della Cassazione n. 6653) ha chiarito che il titolare della servitù non partecipa al condominio agli effetti degli articoli 1117 (parti comuni dell'edificio), 1102 (uso della cosa comune), 1118 (diritti dei partecipanti sulle parti comuni) e 1123 c.c. (ripartizione delle spese). 'esercizio della servitù prediale è invece regolato dalle norme specifiche sulle servitù (artt. 1063 e seguenti c.c.), con particolare riferimento all'art. 1069, comma 3, c.c. per quanto concerne la ripartizione delle spese necessarie per le opere di manutenzione.\\
\hline
\textbf{GPT-4} & Servitù su parti comuni condominiali & Il tema discute della natura delle servitù prediali su parti comuni di un condominio, come l'ingresso e l'androne di un edificio. Si esamina il caso in cui una proprietà esclusiva, esterna al complesso immobiliare condominiale, possa beneficiare di una servitù su beni comuni condominiali. La peculiaritù di tali situazioni è che il godimento di una servitù su parti comuni non comporta automaticamente il riconoscimento dello status di condomino per il titolare della proprietà servente, nè l'acquisizione di diritti o obblighi correlati alla contitolarità delle parti comuni dell'edificio. Tali aspetti sono regolati dagli articoli 1117, 1102, 1118, 1123 del Codice Civile, mentre l'esercizio della servitù e le eventuali spese necessarie sono disciplinati dagli articoli 1063 e seguenti, nonchè dall'articolo 1069, comma 3 del Codice Civile.\\
\hline
\textbf{LLaMA} &La questione relativa alla partecipazione di una proprietà esclusiva ad un condominio e all'esercizio di servitù prediali su beni condominiali, con particolare riferimento agli articoli & Regole e procedure relative alla servitù prediale su beni condominiali e ai rapporti tra proprietà esclusiva e condominio, con particolare riferimento all'applicazione degli articoli 1117, 1118 e 1123 del Codice Civile. \\
\hline
\end{tabular}
\end{table}

\begin{table}[ht]
\centering
\caption{Comparison of model-generated labels and summaries for topic 28}
\label{tab:topic_comparison_10}
\begin{tabular}{|l|p{4cm}|p{7.5cm}|}
\hline
\textbf{Source} & \textbf{Label} & \textbf{Summary} \\
\hline
\textbf{Expert} & Ricorso in Cassazione per errore della prova & Il tema trattato riguarda il travisamento della prova da parte del giudice, che si verifica quando interpreta erroneamente un documento attribuendogli un significato che non corrisponde alla sua reale portata. Secondo la Cassazione, il travisamento della prova può essere denunciato solo quando il giudice commette un errore oggettivo, chiaro, evidente ed incontestabile, tanto da poter essere fatto valere anche in sede di legittimità. Il testo riporta l'esempio di una parte del processo che presenta una fotografia per dimostrare i danni subiti da un'autovettura. Se il giudice afferma erroneamente che l'immagine rappresenta un fiume invece che un'auto, si configura un errore percettivo. Questo tipo di errore può essere oggetto di ricorso in Cassazione, in quanto la prova non è stata correttamente percepita nella sua oggettiva esistenza. Se il giudice riconosce che la fotografia ritrae  un'automobile ma non ne riconosce il corretto valore probatorio, si tratta di una valutazione discrezionale da parte del giudice che può essere erronea ma non contestabile  in sede di legittimità.  \\
\hline
\textbf{Claude} & Il travisamento della prova nel processo civile & Il topic riguarda la dottrina giuridica del "travisamento della prova" nel sistema giudiziario italiano, con particolare riferimento all\'errore evidente di fatto commesso dal giudice nella valutazione delle prove. Si tratta di una questione controversa in dottrina e giurisprudenza, riguardante i casi in cui il giudice percepisce erroneamente il contenuto di un elemento probatorio (come un documento, un contratto, un testamento o una fotografia), attribuendogli un significato oggettivamente diverso da quello reale.La Corte di Cassazione ha adottato nel tempo posizioni contrastanti: mentre una parte della giurisprudenza ha negato rilevanza a tali errori, un'altra corrente ha riconosciuto la censurabilità dell'errore quando non riguarda l'interpretazione della prova ma la sua stessa percezione oggettiva (come nel caso emblematico di scambiare la fotografia di un\'automobile per quella di un fiume). Si distingue tra casi in cui l'errore riguarda un punto controverso tra le parti (non censurabile in Cassazione) e casi in cui l\'errore riguarda la percezione oggettiva di un elemento che non è oggetto di dibattito tra le parti (potenzialmente censurabile).\\
\hline
\textbf{GPT-4} & Errore di fatto e travisamento delle prove nel contesto di contratti e testamenti & Il tema centrale riguarda la complessa problematica dell'errore di percezione e travisamento dei fatti all'interno del contesto giuridico italiano, focalizzandosi sulla distinzione tra contratto e testamento e il loro processo di interpretazione. Vi è una riflessione approfondita sull'errore di fatto, che può sfociare in un travisamento delle prove durante il giudizio. Una parte della dottrina e delle istituzioni segnala l'incompetenza dei giudici nel correggere questi errori quando l'interpretazione del contratto o del testamento è chiara, ma in altri casi, si riconosce la possibilità di correggere l'errore quando riguarda un fatto la cui inesistenza o errata percezione è chiara ed evidente, esempio emblematico è l'interpretazione errata di una fotografia. Inoltre, il dibattito si estende anche al contrasto tra la dottrina e le diverse interpretazioni giurisprudenziali riguardo all'ammissibilità del travisamento come motivo di ricorso in Cassazione. La complessità del dibattito è accentuata dalle diverse prospettive che giudici e giuristi esprimono sulla revisione di giudizi basati su errori evidenti, sia a livello probatorio che sostanziale, nel contesto del sistema legale italiano.\\
\hline
\textbf{LLaMA} & L'analisi del concetto di travisamento della prova nel contesto del diritto italiano, con particolare riferimento alla percezione e alla valutazione dei fatti probatori, e alla possibilità di ricorr & L'analisi del concetto di travisamento della prova nel contesto del diritto italiano, con particolare riferimento alla percezione e all'evidenza dei fatti, e la sua applicazione in casi controversi come la valutazione di prove fotografiche. \\
\hline
\end{tabular}
\end{table}

\section{Conclusion}\label{sec13}

In this paper, we presented a document processing pipeline that transforms raw judicial documents (PDFs) into a dataset optimized for topic modeling, ensuring compliance with privacy regulations for litigants involved in cases published on the Italgiurie website by the Italian Supreme Court.

Our pipeline integrates multiple computer vision techniques, including YOLOv8X for layout analysis, which was fine-tuned on our dataset, achieving solid performance on the test set (mAP50: 0.964, mAP50-95: 0.8). The OCR module was implemented using the YOLOv8X algorithm for line-level text detection (mAP@50-95: 0.9022), combined with TrOCR-small for text recognition, which was also fine-tuned on our dataset, reaching high accuracy (CER: 0.0047, WER: 0.0248).

The results demonstrate that topic modeling techniques such as BERTopic benefit significantly from our pipeline, as it improves text segmentation, enabling the extraction of more diverse and coherent topics (topic diversity: 0.6198, topic coherence ($C_{v}$): 0.663) compared to datasets processed without segmentation.

A thorough BERTopic analysis of the dataset revealed a complex and layered semantic structure within the Italian legal domain. Key legal topics identified include computer fraud, press Defamation, financial intermediation, appeal to the Supreme Court for trial error and appeals related to Article 606 of the Code of Criminal Procedure.

To evaluate topic interpretability, we compared LLM-generated outputs against human expert annotations. Our findings indicate that Claude 3.7 and GPT4o proved to be quite promising in generating summaries (F1 Score: 0.9130 for Claude 3.7 and F1 Score: 0.9122 for GPT) and also at generating a label to the topic (F1 Score: 0.8119 for Claude 3.7 and F1 Score: 0.7995 for GPT). In contrast LLaMA 3.1 (8B) proved to be very mixed, in fact they generated summaries for topics that were semantically adequate but not as rich in detail as those done by Claude 3.7 and GPT4o, and on the topic label generation task it proved to ignore the prompt several times by generating a summary instead of label.

Despite the reliable performance of our pipeline, some limitations should be acknowledged. In particular, the GLiNER anonymization module was used in zero-shot mode due to time constraints in dataset preparation for document layout detection and text recognition. Although our thresholding-based anonymization approach proved effective, it occasionally produced false negatives, which could be mitigated by a more tailored training process.

A potential improvement could involve adapting GLiNER to a customized dataset of Italian legal documents or expanding the dataset by collecting additional judgments from Italgiurie. This enhancement would strengthen the anonymization process, resulting in a more robust and legally compliant pipeline.

Our future works will focus on refining this aspect to further improve the reliability and scalability of our approach to legal document processing and extend this work to other documents of Supreme Court like the ordinances and decrees.

\backmatter

\section*{Declarations}

\bmhead{Funding}
This research was funded by NextGenerationEU PNRR 2022 funds.

\bmhead{Author contribution}
Author contributions according to CRediT taxonomy:
\textbf{Matteo Marulli}: Conceptualization, Methodology, Software, Data Curation, Investigation, Formal Analysis, Writing - Original Draft, Writing - Review \& Editing, Visualization. Led the implementation of the code, Contributed to dataset construction and validation for DLA and OCR datasets , conducted all experiments, wrote the original manuscript, and created the visualizations.\\
\textbf{Glauco Panattoni}: Data Curation, Resources, Validation. Contributed to dataset construction and validation, topic interpretation, provided expertise in LLM selection and preparation, and contributed to defining evaluation metrics for the text generation task.\\
\textbf{Marco Bertini}: Supervision, Project Administration. Supervised the project, provided strategic direction and critical feedback, and reviewed the manuscript.
All authors have read and agreed to the published version of the manuscript.

\bmhead{Data availability}
The datasets generated and analysed during the current study are available from the corresponding author on reasonable request.

\bmhead{Code availability}
The code wrote for this study is available from the corresponding author on his github repository.

\bigskip
\begin{flushleft}%
Editorial Policies for:

\bigskip\noindent
Springer journals and proceedings: \url{https://www.springer.com/gp/editorial-policies}

\bigskip\noindent
Nature Portfolio journals: \url{https://www.nature.com/nature-research/editorial-policies}

\bigskip\noindent
\textit{Scientific Reports}: \url{https://www.nature.com/srep/journal-policies/editorial-policies}

\bigskip\noindent
BMC journals: \url{https://www.biomedcentral.com/getpublished/editorial-policies}
\end{flushleft}








\bibliography{sn-bibliography}

\end{document}